\documentclass[journal]{IEEEtran}
\ifCLASSINFOpdf
\else
\fi
\usepackage[cmex10]{amsmath}
\DeclareMathOperator*{\argmin}{arg\,min} 
\usepackage{bbm}
\usepackage{optidef}
\usepackage{algorithm}
\usepackage{algorithmic}
\usepackage{amssymb,amsfonts}
\usepackage{textcomp}
\usepackage{xcolor}
\usepackage{amsthm}
\usepackage{graphicx}
\usepackage{stfloats}
\usepackage{multirow}
\usepackage{svg}
\usepackage{multicol}

\ifdefined\pdfminorversion\pdfminorversion=7\fi
\usepackage[hidelinks]{hyperref}
\hypersetup{
  pdftitle={Multi-Objective Agent-Based Model Predictive Controller for Plug-and-Play Vehicle Control},
  pdfauthor={Jiaming Zhong; Ladan Khoshnevisan; Shucheng Huang; Mohammad Pirani; Yash Vardhan Pant; Amir Khajepour},
  pdfsubject={Author accepted manuscript; DOI: 10.1109/TITS.2025.3612984}
}

\makeatletter
\newcommand*{\rom}[1]{\expandafter\@slowromancap\romannumeral #1@}
\makeatother

\begin{document}
%
\title{Multi-Objective Agent-Based Model Predictive Controller for Plug-and-Play Vehicle Control}
%
%
%

\author{Jiaming~Zhong,
        Ladan Khoshnevisan,
        Shucheng~Huang,
        Mohammad~Pirani,~\IEEEmembership{Senior Member,~IEEE,}        
        Yash~Vardhan~Pant,        
        and~Amir~Khajepour,~\IEEEmembership{Senior Member,~IEEE}%
\thanks{This work was supported by the Natural Sciences and
Engineering Research Council of CANADA (NSERC) in conducting this research. \textit{(Corresponding author: Jiaming Zhong.)}}%
\thanks{Jiaming Zhong, Ladan Khoshnevisan, Shucheng Huang, and Amir Khajepour are with the Department of Mechanical and Mechatronics Engineering, University of Waterloo, 200 University Ave West, Waterloo ON, N2L3G1 Canada (e-mail: j52zhong@uwaterloo.ca, lkhoshnevisan@uwaterloo.ca, s95huang@uwaterloo.ca, a.khajepour@uwaterloo.ca).}%
\thanks{Mohammad Pirani is with the Department of Mechanical Engineering, University of Ottawa, 75 Laurier Ave East, Ottawa, ON, K1N 6N5 Canada (e-mail: mpirani@uottawa.ca).}%
\thanks{Yash Vardhan Pant is with the Department of Electrical and Computer Engineering, University of Waterloo, 200 University Ave West, Waterloo ON, N2L3G1 Canada (e-mail: yash.pant@uwaterloo.ca).}}

\markboth{AUTHOR ACCEPTED MANUSCRIPT}%
{Zhong \MakeLowercase{\textit{et al.}}: Multi-Objective Agent-Based MPC for Plug-and-Play Vehicle Control}
%



\newsavebox{\AAMnoticebox}
\sbox{\AAMnoticebox}{\parbox{\textwidth}{\footnotesize\raggedright
\textbf{Author accepted manuscript.}
J.~Zhong, L.~Khoshnevisan, S.~Huang, M.~Pirani, Y.~V.~Pant, and A.~Khajepour,
\textquotedblleft{}Multi-Objective Agent-Based Model Predictive Controller for Plug-and-Play Vehicle Control,\textquotedblright{}
\emph{IEEE Transactions on Intelligent Transportation Systems},
vol.~26, no.~12, pp.~22818--22829, December~2025.
The final published version is available at
\href{https://doi.org/10.1109/TITS.2025.3612984}{doi:10.1109/TITS.2025.3612984}.\par\smallskip
\copyright~2025 IEEE. Personal use of this material is permitted.
Permission from IEEE must be obtained for all other uses, in any current or future media,
including reprinting/republishing this material for advertising or promotional purposes,
creating new collective works, for resale or redistribution to servers or lists,
or reuse of any copyrighted component of this work in other works.
}}
\makeatletter
\renewcommand{\@IEEEpubidpullup}{\dimexpr\ht\AAMnoticebox+\dp\AAMnoticebox+\baselineskip\relax}
\makeatother
\IEEEpubid{\usebox{\AAMnoticebox}}

\maketitle

\begin{abstract}
Functional integration is a growing trend in vehicle control, often involving the coordination of multiple controllers to achieve various objectives simultaneously. The need for flexibility and reliability has led to a ``plug-and-play" approach in control system design, which presents challenges for traditional integrated model predictive control (MPC). Agent-based model predictive control (AMPC) has recently emerged as a distributed solution that treats controllers as agents, creating a collaborative framework among them to reach a common goal. However, this approach struggles to manage distributed conflicting objectives when agents are coupled or interdependent. 
{\relax To address this, we propose a novel, practical distributed control scheme called multi-objective AMPC, which adapts the alternating direction method of multipliers (ADMM) into a general control strategy that approximates global optimization while decoupling objectives. }
We systematically develop three formulations that maintain convergence while addressing control regularization and inequality constraints, applying them to complex vehicle control systems for the first time. The proposed method has been tested on two vehicle control scenarios with a multi-objective topology. Different formulations are compared
through simulations, and the most computationally efficient one was implemented on an electric vehicle for real-world evaluations. 
{\relax The results demonstrate that the proposed multi-objective AMPC can converge approximately to the same global optimum as integrated MPC with greater flexibility and the potential to reduce computational costs.}
\end{abstract}

\begin{IEEEkeywords}
Agent-based model predictive control (AMPC), alternating direction method of multipliers (ADMM), distributed vehicle control, multi-agent system.
\end{IEEEkeywords}

%
\IEEEpeerreviewmaketitle

\newpage
\section{Introduction}
%
%
%

\subsection{Motivation}

\IEEEPARstart{V}{ehicle} {\relax systems now have an ever-increasing demand for a variety of functions in intelligence, safety, and efficiency. As a result, more mechatronic systems are being integrated to enhance vehicular performance. Significant efforts are being invested in developing a sophisticated control system to manage all on-board mechatronic systems effectively. However, the demand for flexible configurations requires a ``plug-and-play" fashion for the control system design, which is challenging for traditional integrated model predictive control (MPC). {\relax The ``all-in-one” form in the integrated MPCs may also face challenges like computational burdens and difficulty of maintenance \cite{mazzilli2021integrated}\cite{khalatbarisoltani2023integrating}. }
{\relax In contrast, a distributed architecture is more in line with today's modular design in many areas, including the vehicle industry. }
Therefore, decoupling complex vehicle control systems into a distributed topology emerges as a natural solution \cite{tang2021agent}.}

\IEEEpubidadjcol
{\relax Multi-agent control systems (MACSs) usually consist of a network of agents that cooperate based on an interacting topology,} which is widely used in applications of cooperative planning \cite{van2016online}, formation control \cite{xin2023model}, and traffic control \cite{10125035}. Although ``agents" often represent individual vehicles or robots, MACSs have also been applied to control networks within a vehicle system where multiple controllers are regarded as connected agents. {\relax As more controllers are incorporated into vehicle systems, decomposing the traditional integrated control architecture into a distributed MACS scheme can improve flexibility and scalability.}

\begin{figure}[!t]
\centering
\includegraphics[width=3.49in]{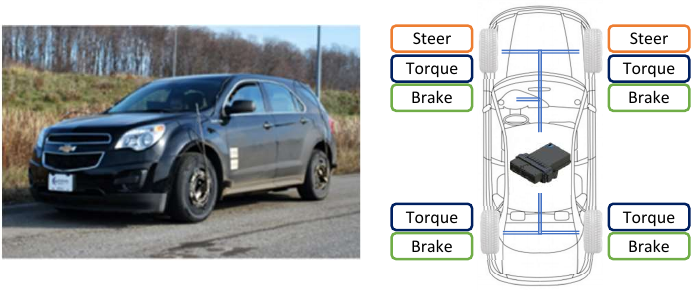}
\caption{{\relax Test vehicle with independent wheel control actuators. Distributed controllers, such as torque-vectoring and differential braking, are coordinated as connected agents to achieve multiple objectives. We use it for experimental evaluation of our proposed methods in section \ref{Examples}.}}
\setlength{\tabcolsep}{3pt}
\label{exp_vehicle}
\end{figure}

Fig. \ref{exp_vehicle} shows a typical MACS example used in this work for vehicle control. Each wheel corner can be controlled independently. Controllers, such as torque-vectoring and differential braking, are treated as agents and coordinated to achieve multiple objectives, such as cruise control and stability control. 

{\relax An essential requirement of distributed control schemes for MACSs is to obtain approximate results as the integrated one - the global optimum.} 
This requirement becomes challenging for many previously proposed distributed MPC (DMPC) solutions when subsystems have conflict objectives on shared agents. 
As one of the DMPCs, the recently proposed agent-based MPC (AMPC) scheme \cite{tang2020wheel} has shown the ability to converge to the centralized MPC using an iterative consensus algorithm.
In this approach, however, the system is decomposed from the agent's perspective, and only one objective is considered, which does not apply to some practical complex systems.
To this end, this paper aims to systematically propose a multi-objective AMPC scheme tailored from the alternating direction method of multipliers (ADMM), in which objectives are distributed with a corresponding agent group. 
{\relax The global optimum should be approximated iteratively through information exchange between agent groups.}
The proposed scheme will have excellent flexibility since objectives, i.e., vehicle control features such as speed cruise control or yaw stability control, can be designed in a ``plug-and-play" fashion. Moreover, efficiency improvement will be a potential advantage.

\subsection{Literature Review}

{\relax Early application of DMPC can be found in \cite{elliott2008model}, 
where the interconnections between different local MPCs are assumed to be weak and are neglected \cite{liu2024distributed}.}
For modern DMPC, communication and information exchange in a network are needed. Unlike the non-cooperative DMPC {\relax \cite{zhang2025efficient}}, which usually leads to the Nash equilibrium as local MPCs are selfish, cooperative DMPC optimizes the same global cost function at each local controller \cite{6199971}.
Commonly used forms in wider applications of cooperative DMPC include sequential DMPC \cite{liu2010sequential} and iterative DMPC \cite{maestre2011distributed}. 
{\relax However, approximating to the global optimum as the corresponding integrated formulation is challenging. Many DMPCs used in some formation control tasks of unmanned aerial vehicle (UAV) \cite{8802250} and vehicle platooning \cite{7546918} cannot approximate the global optimally.}

{\relax The recently proposed agent-based AMPC \cite{zhong2024learning} is a type of cooperative DMPC based on a consensus mechanism and has been proven in vehicle control systems to converge to a centralized MPC. The main idea of AMPC is that each controller is individually modeled as an interactive agent assigned with a local control task.} 
Agents could be connected in a flexible plug-and-play fashion while emulating centralized MPC performance. 
{\relax However, this method is limited to the many complex MACSs in practice with a ``multi-input-multi-output" topology, where multiple objectives need to be decomposed, and there are couplings or shared agents between those distributed objectives.}

As one of the distributed optimization algorithms, ADMM facilitates the solution of the consensus problem for MACSs \cite{9632418}. It has found a large number of applications in areas like wireless networks \cite{joshi2013distributed}, machine learning \cite{forero2010consensus}, and parallel computation \cite{schubiger2020gpu}. 
{\relax Thanks to the strong convergence and ability to decompose, there have been some studies tailoring ADMM to DMPC applications such as cooperative planning \cite{zhou2023distributed}, formation control \cite{7572966}, and smart power grid \cite{shi2021distributed}.}

{\relax However, as far as the authors of this paper know, there is no systematic discussion or practical application of the ADMM method to complex vehicle control systems. }

\subsection{Contributions}

{\relax This paper proposed the multi-objective AMPC scheme as a generic and practical distributed solution for MACSs and applied it to vehicle control systems. The main contributions of this work are summarized as follows:}

\begin{itemize}
    \item A novel distributed control scheme - multi-objective AMPC, tailored from the ADMM, has been proposed. Three practical formulations that can handle the control regularization and inequality constraints are systematically demonstrated and compared through simulations. 
    \item The proposed formulations have been implemented on an electric vehicle for real-time validation. {\relax The results demonstrate that the proposed multi-objective AMPC can converge approximately to the same global optimum as integrated MPC while exhibiting potentially superior computational efficiency in vehicle control applications.}
\end{itemize}

The paper is structured as follows. Section \ref{Problem Demonstration} demonstrates the MACS topology and the corresponding integrated MPC solution used in this paper. Then, several multi-objective AMPC formulations tailored from ADMM are given in Section \ref{Formulations}. Two MACS examples on vehicle control are introduced in Section \ref{Vehicle Examples}. Real-time experiments and results on two examples are demonstrated in Section \ref{Examples}. Conclusions are summarized in Section \ref{Conclusion}.


\section{{\relax Multi-objective MACS}} \label{Problem Demonstration}

{\relax 
\subsection{Notation}

In this paper, the term ``agent" refers to a controller in the MACS.
Subscripts of index $j$ and index in parentheses $(i)$ are used for notations referring to the objective $O_j$ and the agent $A_{(i)}$, respectively. Superscripts of index in parentheses $(k)$ and index in angle brackets $\langle q \rangle$ are used for the prediction step in the horizon and iterations in one time step, respectively.
Unless stated otherwise, iteration markers $\langle q \rangle$ are omitted from all variables in this paper. Superscripts without parentheses or angle brackets denote powers as commonly used.

{\relax We define the set $S_A = \{A_{(i)}\}$ and $S_O = \{O_j\}$ to represent all agents and objectives in a MACS, respectively.
Each agent $A_{(i)}$ generates its own control actions denoted as a vector $U_{(i)} \in \mathbb{R}^{n_{u(i)}}$. Thus, the overall control inputs $U = [U_{(i)}^T | i\in S_A]^T \in \mathbb{R}^{n_{u}}$ is the concatenation of all agents' control actions with $n_u = \sum n_{u(i)}$.
We also define the set $S_j$ representing the index set of agents who have contributions on the objective  $O_j$.}
The vector of control actions that contribute to objective $O_j$ is defined as $U_j = [U_{(i)}^T | i\in S_j ]^T \in \mathbb{R}^{n_{uj}}$, where $n_{uj} = \sum_{i\in S_j} n_{u(i)}$. $U_j$ is a projection from $U$. 

Similarly, we use $Z \in \mathbb{R}^{n_{z}}$ to represent the common global variable in ADMM and $Z_j \in \mathbb{R}^{n_{zj}}$ to represent the local projection from $Z$ for the objective $O_j$. The values of dimensions $n_{z}$ and $n_{zj}$ may differ depending on the distributed formulation in Section \ref{Formulations}.}

{\relax 
Over-lines are used to represent sequences along the prediction horizon: $\bar{U} = [{(U^{(k)})}^T | k=0, \ldots, N_p-1]^T \in \mathbb{R}^{n_u \cdot N_p}$, $\bar{U}_j = [{(U_j^{(k)})}^T | k=0, \ldots, N_p-1]^T \in \mathbb{R}^{n_{uj} \cdot N_p}$, $\bar{Z} = [{(Z^{(k)})}^T | k=0, \ldots, Z_p-1]^T \in \mathbb{R}^{n_z \cdot N_p}$, and $\bar{Z}_j = [{(Z_j^{(k)})}^T | k=0, \ldots, N_p-1]^T\in \mathbb{R}^{n_{zj} \cdot N_p}$. 
Besides, we use $P_j \in \mathbb{R}^{n_{zj} \cdot N_p \times n_z \cdot N_p}$ as the projection matrix for $\bar{Z}_j = P_j \cdot \bar{Z}$.}

A Boolean value $G_{j(i)}=1$ is used to represent that the agent $A_{(i)}$ contributes to (or has influence on) the objective $O_j$. Otherwise, there is no relationship between $A_{(i)}$ and $O_j$. The value of $G_{j(i)}$ can usually be obtained by analyzing the system model, such as the partial differential equation of the objective on the agent. We further define:
\begin{itemize}
    \item \textit{Agent group for $O_j$}: $S_j = \{ i | G_{j(i)}=1\}$, the index set of agents that contribute to the objective $O_j$.
    \item \textit{Contribution set for $A_{(i)}$}: $S_{(i)} = \{ j | G_{j(i)}=1\}$, the index set of objectives that the agent $A_{(i)}$ contributes to.
\end{itemize}

\begin{figure}[!t]
\centering
\includegraphics[width=3.1in]{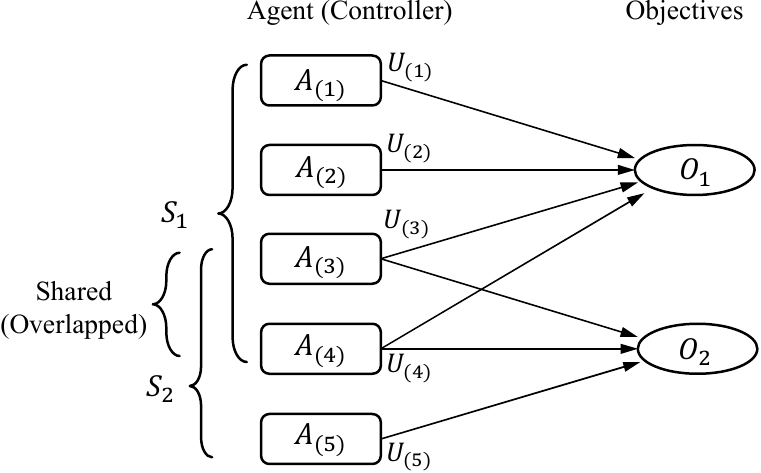}
\caption{{\relax Example of a MACS with five agents and two objectives. The two objectives share two agents. The arrow from $A_{(i)}$ to $O_j$ represents $G_{j(i)}=1$. Thus, for objectives: $S_1 = \{1, 2, 3, 4\}$, $S_2 = \{3, 4, 5\}$. $U_1 = [U_{(1)}^T, U_{(2)}^T, U_{(3)}^T, U_{(4)}^T]^T$, and $U_2 = [U_{(3)}^T, U_{(4)}^T, U_{(5)}^T]^T$. For agents: $S_{(1)} = \{1\}$, $S_{(2)} = \{1\}$, $S_{(3)} = \{1, 2\}$, $S_{(4)} = \{1, 2\}$, and $S_{(5)} = \{2\}$.}}
\setlength{\tabcolsep}{3pt}
\label{overall_structure}
\end{figure}

{\relax A MACS could be easily decoupled into fully independent subsystems for each objective if there is no shared agent ($S_j \cap S_{j'} = \emptyset$, $\forall j \neq j'$).} However, this is often too idealistic in practice. To this end, the multi-objective AMPC proposed in this study aims to decompose the MACS with shared agents.
An example of this type of MACSs with two objectives and several agents is shown in Fig. \ref{overall_structure}.

The $n \times n$ identity matrix and the $n$-dimensional column vector with all elements taking the value 0 are denoted by $I_n$ and $\mathbf{0}_n$.
We define $\|x\|^2_2 = x^Tx$ and $\|x\|^2_Q = x^TQx$ for positive semi-definite matrix $Q$. 
{\relax $L_+(x)$ is an indicator function of $x$ that $L_+(x) = 0$ for $x \geq 0$ and $L_+(x) = +\infty$ otherwise.}

{\relax 
\subsection{Integrated MPC}}

{\relax 
First, we establish the integrated MPC for the MACS as a reference for distributed methods. 
For multi-objective optimization problems, the most common approach is constructing an integrated cost function in the form of weighted summation.} Thus, the discrete integrated MPC can be written as,
\begin{mini}|s|[0]
    {{\relax \bar{U}}}
    {{\relax J = \sum_{k=1}^{N_p} \sum_{j=1}^{|S_O|} h_j(X^{(k)}) + \sum_{k=1}^{N_p} g(U^{(k-1)}) }}
    {\label{eq.integMPC}}{}
    \addConstraint{X^{(k+1)} = f_d (X^{(k)}, U^{(k)}) , k \in \{ 0, \ldots, N_p-1 \}}
    \addConstraint{U^{(k)} \in \Omega^{(k)} , k \in \{ 0, \ldots, N_p-1 \}}
\end{mini}
{\relax where, $J$ is the cost funciton. $X\in \mathbb{R}^{n_x}$ is the system state and $X^{(0)}$ is the current state from measurement or estimation, $N_p$ is the length of prediction horizon, $\Omega$ is the control constraint set, $h_j$ is the state penalty for objective $O_j$, $g$ is the regularization function for $U$, $f_d$ is the discrete system equation.}

{\relax For many real-time control systems, such as vehicles, it is common to approximate nonlinear MPC as linear MPC to solve for higher computational efficiency.}
Therefore, this paper focuses on the optimization with quadratic cost functions, linear systems equations, and linear inequality constraints:
\begin{equation} \label{eq_linear}
\relax 
\begin{aligned}
    h_j(X^{(k)}) :=& {\| C_j (X^{(k)} - X_{ref}^{(k)}) \|}_{Q_j}^2, \\
    g(U^{(k)}) :=& \sum_{i=1}^{|S_A|} g_{(i)}(U_{(i)}^{(k)}) = \sum_{i=1}^{|S_A|} {\| U_{(i)}^{(k)} \|}_{R_{(i)}}^2 , \\
    f_d(X^{(k)}, U_i^{(k)}) :=& A_d X^{(k)} + \sum_{i=1}^{|S_A|} B_{d{(i)}} U_i^{(k)} + W_d^{(k)},\\
    \Omega^{(k)}: =& A_{in} U^{(k)} \leq b_{in}.
\end{aligned}
\end{equation}
where $A_d$, $B_{d(i)}$, and $W_d$ are the system matrix, control matrix, and disturbance term, respectively. $C_j$ is the observation matrix for objective $O_j$. Positive-definite $Q_j$ and semi-positive definite $R_{(i)}$ are penalty weights for tracking error and control action, respectively. $X_{ref}$ denotes the desired tracking reference. 
$A_{in}\in \mathbb{R}^{n_c \times n_u}$ and $b_{in}\in \mathbb{R}^{n_c \times 1}$ are the coefficient matrix and constant vector of the linear inequality constraints, respectively, where $n_c$ is the dimension of constraints.
Based on (\ref{eq_linear}), the integrated MPC in (\ref{eq.integMPC}) should be convex, and its solution is the global optimum.

\section{Multi-objective Agent-based MPC} \label{Formulations}

\subsection{Basic Algorithm of Multi-objective AMPC}

The basic idea of the proposed multi-objective AMPC is to tailor the ADMM to fit MACSs with $ S_{S} \neq \emptyset$. 
ADMM is a well-known method for solving optimization problems for multi-agent systems. It approximates the global optimum using iterative local optimizations on the augmented Lagrangian equation. 
{\relax Through the dual ascent method, it is proven to achieve residual convergence, objective convergence, and dual convergence \cite{boyd2011distributed}, \cite{rostami2017admm}, \cite{ghadimi2014optimal}.} 

In particular, the consensus formulation proposed in \cite{boyd2011distributed} was found to be fitted for distributed control systems. Assume the cost function in (\ref{eq.integMPC}) can be broken down into local costs as {\relax $J_j(\bar{U}_j)$} and ignore the inequality constraints $\Omega$, then we will have a basic multi-objective AMPC based on the standard consensus ADMM problem as,
\begin{mini}|s|[0]
    {\bar{U}_j}
    {{\relax J = \sum_{j=1}^{|S_O|} J_j(\bar{U}_j)}} 
    {\label{eq.ConsensusADMM}}{}
    \addConstraint{{\relax \bar{U}_j - \bar{Z}_j = \mathbf{0}_{n_{zj} \cdot N_p}}}
\end{mini}
where {\relax $J_j$} are the cost functions of local optimizations on $\bar{U}_j$. There is a one-to-one correspondence between every element in $\bar{U}$ and $\bar{Z}$, and so do $\bar{U}_j$ and $\bar{Z}_j$. Thus, $n_z = n_u$ and  $n_{zj} = n_{uj}$. The constraint in (\ref{eq.ConsensusADMM}) implies the requirement that all local variables in $\bar{U}$ shall converge to the same global variable $\bar{Z}$ in the end since $\bar{Z}_j$ are projections from $\bar{Z}$. 

{\relax 
The augmented Lagrangian for (\ref{eq.ConsensusADMM}) is constructed as,
\begin{equation} \label{eq.Lagrangian}
\relax 
\begin{aligned}
    L_\rho (\bar{U}_1, \ldots, \bar{U}_J, \bar{Z}, \gamma_1,  \ldots, \gamma_J) &\\
    = \sum_{j=1}^{|S_O|} \Bigl[ J_j(\bar{U}_j) + \gamma_j^T \bar{U}_j & + \frac{\rho_j}{2}{\| \bar{U}_j - \bar{Z}_j \|}_{2}^2 \Bigr]
\end{aligned}
\end{equation}
where $\gamma_j \in \mathbb{R}^{n_{zj} \cdot N_p}$ is the Lagrange multiplier as the dual variable, and $\rho_j \in \mathbb{R} > 0$ is a penalty parameter.}

Algorithm \ref{alg:1} shows the basic multi-objective AMPC algorithm for (\ref{eq.ConsensusADMM}) based on the consensus ADMM. 
It is guaranteed by ADMM that if the problem (\ref{eq.ConsensusADMM}) has a global optimum, which is a saddle point ($\bar{U}$, $\bar{Z}$, $\gamma$), (i) the objective evaluated at the primal variables ($\bar{U}$, $\bar{Z}$) converges to its optimal value, (ii) the primal residual ($\bar{U} - \bar{Z}_j$) converge to zero, and (iii) the dual variable $\gamma$ will converge to a saddle point \cite{pan2024asynchronous}.

\begin{algorithm}
    \caption{Basic Algorithm of Multi-objective AMPC}
    \label{alg:1}    
    \begin{algorithmic}[1]
        \STATE {\relax \textbf{Input:} local cost functions $J_j$ and $\rho_j$}
        \STATE Local initialization $\gamma_j = \mathbf{0}_{n_{zj} \cdot N_p}$ and $\bar{Z}_j= \mathbf{0}_{n_{zj} \cdot N_p}$
        \STATE Global initialization $\bar{Z} = \mathbf{0}_{n_z \cdot N_p}$
        \REPEAT
            \STATE \label{step1} \textbf{Local Update} on each objective: 
            $\bar{U}_j = \displaystyle\argmin_{\bar{U}_j} L_{\rho}$ 
            \STATE Broadcast $\bar{U}_j$
            \STATE \label{step2} \textbf{Global Update} on
            $\bar{Z} = \displaystyle\argmin_{\bar{Z}} L_{\rho}$ 
            \STATE Project $\bar{Z}$ to update $\bar{Z}_j$
            \STATE \label{step3} \textbf{Dual Update} on each Lagrange multiplier:\\
            $\gamma_j = \gamma_j + \rho_j (\bar{U}_j - \bar{Z}_j)$
        \UNTIL{{\relax convergence achieved through (\ref{eq.Convergence}) for all objectives}}
        \STATE \textbf{Output:} The first step of the control variable: $\bar{U}^{(0)}$
    \end{algorithmic}    
\end{algorithm}

The convergence conditions of objective $O_j$ are,
\begin{equation} \label{eq.Convergence}
\begin{aligned}
    & \text{Primal:} ~ \|r_j^{\langle q \rangle}\|_2^2 = {\| \bar{U}_j^{\langle q \rangle} - \bar{Z}_j^{\langle q \rangle} \|}_2^2 \leq \epsilon_r \text{, AND}\\
    & \text{Dual:} ~ \|s_j^{\langle q \rangle}\|_2^2 = {\| \rho (\bar{Z}_j^{\langle q \rangle} - \bar{Z}_j^{\langle q-1 \rangle} )\|}_2^2 \leq \epsilon_s 
\end{aligned}
\end{equation}
where, $r_j^{\langle q \rangle}$ and $s_j^{\langle q \rangle}$ are the primal and dual residuals at the iteration $q$, respectively. $\epsilon_r > 0$ and $\epsilon_s > 0$ are the thresholds.

{\relax However, the above basic algorithm neither explains how to decompose the cost function $J$ in (1) into an equivalent distributed form as $J_j$ in (3) nor does it contain any inequality constraints, which are crucial in establishing a generic distributed MPC for practical MACS.} 
Therefore, three feasible formulations that could meet the requirement are discussed in Section \ref{form 1} - \ref{form 3}, {\relax where the key steps of local update, global update, and dual update are given}.

\subsection{Formulation 1: Fully Distributed} \label{form 1}

{\relax As the name describes, this formulation distributes every component related to the local variable $U_j$ into the local optimization, the step-\ref{step1} in Algorithm \ref{alg:1}.} Therefore, both the state penalty $h_j$ and the control regularization $g_{(i)}$ are distributed into the local cost function {\relax $J_j$}:
\begin{equation} \label{eq.F1_obj}
\relax 
    J_j(\bar{U}_j) = \sum_{k=1}^{N_p} \Bigl(h_j(X^{(k)}) + \frac{1}{m_{(i)}} \sum_{i\in S_j} g_{(i)}(U_{(i)}^{(k-1)}) \Bigr)
\end{equation}
where $m_{(i)} = |S_{(i)}|$ is the number of objectives that the agent $A_{(i)}$ is contributing through $U_{(i)}$. Therefore $m_{(i)}$ should be greater than 1 if $A_{(i)}$ is a shared agent, otherwise $m_{(i)} = 1$. Therefore, {\relax $J = \sum J_j$ is satisfied.}
Besides, the inequality constraints on $U_j$ should also be considered separately in each local optimization. 

\textbf{{\relax Local update:}} Thus, the optimization in (\ref{eq_linear}) for each local update in algorithm \ref{alg:1} can be written as,
{\relax 
\begin{mini}|s|[0]
    {\bar{U}_j}
    {\begin{aligned}\sum_{k=1}^{N_p} 
    \Bigl({\| C_j (X^{(k)} - X_{ref}^{(k)}) \|}_{Q_j}^2\Bigr) +  {\| \bar{U}_j \|}_{\bar{R}_j}^2 \\
    + \gamma_j^T \bar{U}_j + \frac{\rho_j}{2} {\| \bar{U}_j - \bar{Z}_j \|}_2^2 
    \end{aligned}} 
    {\label{eq.F1_primal}}{}
    \addConstraint{C_j X^{(k+1)} = C_j (A_d X^{(k)} + B_{d,j} U_j^{(k)} + W_d^{(k)}) }
    \addConstraint{U_j^{(k)} \in \Omega_j^{(k)} := A_{in,j} U_j^{(k)} + B_{in,j} \tilde{U}_j^{(k)} \leq b_{in,j}}
\end{mini}
where, 
\begin{equation} \label{eq.F1_1}
\begin{aligned}
    &\bar{R}_j = \text{diag}\Bigl( [\underbrace{R_j, \dots, R_j}_{N_p}] \Bigr), 
    R_j = \text{diag}\Bigl(\Bigl[\frac{R_{(i)}}{m_{(i)}} \Big| i \in S_j \Bigr]\Bigr), \\
    &B_{in,j} = [B_{d{(i)}} | i \in S_j], 
    \tilde{U}_j^{(k)\langle q \rangle} = [{(U_{(i)}^{(k)\langle q-1 \rangle})}^T | i \notin S_j]^T.
\end{aligned}
\end{equation}}

{\relax In Equation (\ref{eq.F1_primal}), $A_{in,j}\in \mathbb{R}^{n_{cj} \times n_{uj}}$, $B_{in,j}\in \mathbb{R}^{n_{cj} \times (n_u-n_{uj})}$ and $b_{in,j}\in \mathbb{R}^{n_{cj}}$ are the matrices and vector of the inequality constraints for $U_j$, respectively, where $n_{cj}$ is the dimension of constraints related to $U_j$. 
{\relax In this \textit{Fully Distributed} formulation, inequality constraints will be considered in each local optimization. $\tilde{U}_j^{(k)}$ is the control actions from the agents that are not included in the agent group $S_j$ and are generated from the previous iteration. In this way, as each local optimization exchanges information through $\tilde{U}_j^{(k)}$ during iterations, each local inequality constraint in (\ref{eq.F1_primal}) will approach the global inequality in the original integrated form in (\ref{eq_linear}).}
$A_{in,j}$, $B_{in,j}$, and $b_{in,j}$ are properly calculated from $A_{in}$ and $b_{in}$ to make the constraint in (\ref{eq.F1_primal}) be equivalent to the constrain in (\ref{eq_linear}) on $U_j$.}


{\relax \textbf{Global update:} After the local update, each control action $U_{(i)}^{(k)}$ in $\bar{U}$ (or the corresponding $Z_{(i)}^{(k)}$ in $\bar{Z}$) will receive $m_{(i)}$ distributed updates.} The optimization on the global variable $\bar{Z}$ in algorithm \ref{alg:1} can be easily simplified as a global average for each $i$-th element as \cite{boyd2011distributed},
{\relax 
\begin{equation} \label{eq.F1_dual}
    Z_{(i)}^{(k)} = \frac{1}{m_{(i)}} \sum_{j \in S_{(i)}} U_{j(i)}^{(k)}
\end{equation}
where $U_{j(i)}^{(k)}$ is the control action for agent $A_{(i)}$ updated from the local optimization for $O_j$ at the $k$-th step.}

{\relax 
\textbf{{\relax Dual update:}} Keep the same update on $\gamma_j$ as Algorithm \ref{alg:1}.}

{\relax 
The global and dual updates are simple and computationally efficient in this formulation. However, the optimization in the local update is relatively complex because the regularization of the control action is considered in each local node. }

\subsection{Formulation 2: Virtual Center Node} \label{form 2}

In this formulation, the regularization of the control action is handled centrally, while only state penalty $h_j$ is distributed in the local cost function {\relax $J_j$}. It is called ``virtual" because the centralized regularization can be computed on a physical center node or replicated on each local node. Hence, the integrated cost function in (\ref{eq.integMPC}) can be written as,
\begin{equation} \label{eq.ConsensusADMM_F2}
\relax 
    J = G(\bar{Z}) + \sum_{j=1}^{|S_O|} J_j(\bar{U}_j)
\end{equation}
where,
\begin{equation} \label{eq.ConsensusADMM_F2_1}
\relax 
\begin{aligned}
    J_j(\bar{U}_j) = \sum_{k=1}^{N_p} h_j(X^{(k)}),~~ G(\bar{Z}) = \sum_{k=1}^{N_p} g(Z^{(k-1)}).
\end{aligned}
\end{equation}

The function $G(\bar{Z})$ could handle the regularization for all the global control variables in a (virtual) center node.
Then, the augmented Lagrangian is rewritten as,
\begin{equation} \label{eq.Lagrangian_F2}
\begin{aligned}
    L_\rho (\bar{U}_1, \ldots, \bar{U}_J, \bar{Z}, \gamma_1,  \ldots, \gamma_J) &\\
    = G(\bar{Z}) + \sum_{j=1}^{|S_O|} \Bigl[ f_j(\bar{U}_j) +& \gamma_j^T \bar{U}_j  + \frac{\rho_j}{2}{\| \bar{U}_j - \bar{Z}_j \|}_{2}^2 \Bigr]
\end{aligned}
\end{equation}

\textbf{Local update:} {\relax Similar to the \textit{Fully Distributed} formulation, inequality constraints should also be considered separately in each local optimization. Each local optimization exchanges information through $\tilde{U}_j^{(k)}$ during iterations to approximate the original global inequality constraint.} The optimization for each local update in algorithm \ref{alg:1} can be written as,
{\relax 
\begin{mini}|s|[0]
    {\bar{U}^j}
    {\begin{aligned}
    \sum_{k=1}^{N_p} 
    \Bigl({\| C_j (X^{(k)} - X_{ref}^{(k)}) \|}_{Q_j}^2 \Bigr) 
    + {\gamma_j}^T \bar{U}_j \\ + \frac{\rho_j}{2}{\|\bar{U}_j - \bar{Z}_j \|}_2^2 
    \end{aligned}} 
    {\label{eq.F2_primal}}{}
    \addConstraint{C_j X^{(k+1)} = C_j (A_d X^{(k)} + B_{d,j}^j U_j^{(k)} + W_d^{(k)}) }
    \addConstraint{U_j^{(k)} \in \Omega_j^{(k)} := A_{in,j} U_j^{(k)} + B_{in,j} \tilde{U}_j^{(k)} \leq b_{in,j}}
\end{mini}
where $A_{in,j}$, $B_{in,j}$, $b_{in,j}$, and $\tilde{U}_j^{(k)}$ have the same definition as in (\ref{eq.F1_primal}) and (\ref{eq.F1_1}).}

{\relax \textbf{Global update:} The regularization function on control actions is handled directly on the global variable $Z$ instead of the local variable $U_j$. Therefore, the global update in algorithm \ref{alg:1} can be derived as,
\begin{equation} \label{eq.F2_dual}
Z_{(i)}^{(k)} = \frac{\sum_{j \in S_{(i)}} (\gamma_{j(i)}^{(k)} + \rho_j \cdot U_{j(i)}^{(k)})}{2 R_{(i)} + \sum_{j \in S_{(i)}} \rho_j }
\end{equation}
where $\gamma_{j(i)}^{(k)}$ is the $(k\cdot N_p + i)$-th element in $\gamma_j$. The regularization penalty $R_{(i)}$ is shown as a part of the denominator: the $Z_{(i)}^{(k)}$ would be close to 0 if $R_{(i)}$ holds a very big value, while the $Z_{(i)}^{(k)}$ would be close to (\ref{eq.F1_dual}) if $R_{(i)}$ holds a very small value since $\gamma_j$ would be zeros after the first iteration.

\textbf{Dual update:} Keep the same update on $\gamma_j$ as Algorithm \ref{alg:1}.

The global update using an explicit average expression in this formulation is still efficient, while the local update is a bit simpler than the \textit{Fully Distributed} formulation.}

\subsection{Formulation 3: Local Explicit} \label{form 3}

{\relax 
The biggest difference, as well as the main advantage, of this formulation is converting the inequality constraints into indicator functions showing in the cost functions, allowing local updates to be solved explicitly through simple matrix operations.}
To achieve this, the common global variables $Z$ will no longer correspond to control actions $U$. Instead, inspired by \cite{ghadimi2014optimal}, the global variable $Z$ will be redefined as a slack variable to transform the inequality constraints into equality constraints. 
Hence, the integrated MPC in (\ref{eq.integMPC}) can be rewritten as, 
\begin{mini}|s|[0] 
    {\bar{U}}
    {\relax J = L_+(\bar{Z}) + \sum_{j=1}^{|S_O|} J_j(\bar{U}_j) ~~~~~~~~~~~~~~~~~~}
    {\label{eq.integMPC_F3}}{}
    \addConstraint{X^{(k+1)} = f_d (X^{(k)}, U^{(k)})}
    \addConstraint{U^{(k)} \in \Omega^{(k)}:= A_{in} U^{(k)} - b_{in} + Z^{(k)} = \mathbf{0}_{n_{z}}}
\end{mini}
where {\relax $J_j(\bar{U}_j)$} has the same expression as the \textit{Fully Distributed} formulation in (\ref{eq.F1_obj}). $L_+$ is an indicator penalty function that puts an infinite penalty on negative components of the slack variable $\bar{Z}_j$, which makes it equivalent to the original optimization problem. 
It is worth noting that the dimension of $Z$ in this formulation is no longer the same as $U$. Instead, the dimension of $Z$ is the same as that of the constraint, that is, $n_z = n_c$ and  $n_{zj} = n_{cj}$.
Then, the augmented Lagrangian is rewritten as,
\begin{equation} \label{eq.Lagrangian_F3}
\relax 
\begin{aligned}
    L_\rho (\bar{U}_1, \ldots, \bar{U}_J, \bar{Z}, \gamma_1,  \ldots, \gamma_J) &\\
    = L_+(\bar{Z}) + \sum_{j=1}^{|S_O|} \Bigl[ J_j(\bar{U}_j) +& \gamma_j^T \bar{U}_j  + \frac{\rho_j}{2}{\| \bar{U}_j - \bar{Z}_j \|}_{2}^2 \Bigr]
\end{aligned}
\end{equation}

It is worth mentioning that, since the indicator function $L_+(\bar{Z})$ is not related to $\bar{U}_j$, {\relax the indicator function $L_+(\bar{Z})$ and the equality constraint $U^{(k)} \in \Omega^{(k)}$ can be considered in the global update instead of the local update, to simplify the optimization of each local update. }

\textbf{{\relax Local update:}} The $j$-th local update can be written as,
{\relax 
\begin{mini}|s|[0]
    {\bar{U}_j}
    {\begin{aligned}\sum_{k=1}^{N_p} 
    \Bigl({\| C_j (X^{(k)} - X_{ref}^{(k)}) \|}_{Q_j}^2\Bigr) +  {\| \bar{U}_j \|}_{\bar{R}_j}^2 \\
    + \gamma_j^T \bar{U}_j + \frac{\rho_j}{2} {\| \bar{U}_j - \bar{Z}_j \|}_2^2
    \end{aligned}} 
    {\label{eq.ConsensusADMM_F3}}{}
    \addConstraint{C_j X^{(k+1)} = C_j (A_d X^{(k)} + B_{d,j} U_j^{(k)} + W_d^{(k)}) }
\end{mini}

Therefore, the local update in (\ref{eq.ConsensusADMM_F3}) can be simply solved in an explicit expression as,
\begin{equation} \label{eq.F3_primal}
\begin{aligned}
    \bar{U}_j =&-{(H_{j} + \rho_j \bar{A}_{in,j}^{T} \bar{A}_{in,j})}^{-1} \\ 
    &~~ \cdot \Big[ E_{j} + \rho_{j} \bar{A}_{in,j}^{T}  
    (\bar{Z}_{j} + \frac{1}{\rho^j} \gamma^j - \bar{b}_{in,j} ) \Big]
\end{aligned}    
\end{equation}
{\relax where, 
\begin{equation} \label{eq.F3_primal_1}
\begin{aligned}
    H_j &= 2 \Bigl(Y_{u,j}^T \bar{Q}_j Y_{u,j} + \bar{R}_j \Bigr),\\
    E_j &= 2 Y_{u,j}^T \bar{Q}_j \Bigl( Y_{x,j} X^{(0)} + Y_{w,j} \bar{W_j} - \bar{X}_{ref,j}  \Bigr),\\
    Y_{u,j} &= \begin{bmatrix}
                C_j A_d^{0} B_{d,j}       & \cdots      & \mathbf{0} \\
                \vdots & \ddots & \mathbf{0} \\
                C_j A_d^{N_p-1} B_{d,j}  & \cdots & C_j A_d^{0} B_{d,j}
                \end{bmatrix},\\
    Y_{x,j} &= {\Bigl[{(C_j A_d^{p})}^T ~|~ p=1, \ldots, N_p\Bigr]}^T,\\
    Y_{w,j} &= \begin{bmatrix}
                C_j A_d^{0}        & \cdots      & \mathbf{0} \\
                \vdots & \ddots & \mathbf{0} \\
                C_j A_d^{N_p-1}   & \cdots & C_j A_d^{0} 
                \end{bmatrix},
\end{aligned}
\end{equation}
and,
\begin{equation} \label{eq.F3_primal_1_1}
\begin{aligned}    
    \bar{Q}_j &= \text{diag}\Bigl( [\underbrace{Q_j, \dots, Q_j}_{N_p}] \Bigr),\\
    \bar{A}_{in,j} &= \text{diag}\Bigl( [\underbrace{{A}_{in,j}, \dots, {A}_{in,j}}_{N_p}] \Bigr), \\
    \bar{b}_{in,j} &= {\Bigl[{(b_{in,j}-B_{in,j} \tilde{U}_j^{(k)})}^T | k=0, \ldots, N_p-1\Bigr]}^T,\\
    \bar{X}_{ref,j} &= {\Bigl[{(C_j X_{ref}^{(k)})}^T ~|~ k=0, \ldots, N_p-1\Bigr]}^T,\\
    \bar{W_j} &= {\Bigl[{(C_j W_d^{(k)})}^T ~|~ k=0, \ldots, N_p-1\Bigr]}^T.         
\end{aligned}
\end{equation}
}


\textbf{Global update:} In this formulation, shared agents lead to shared constraints. 
{\relax Let us define $c = \{1, \ldots, n_c\}$ to index the constraints of $U^{(k)} \in \Omega^{(k)}$.}
Then $Z_{(c)}^{(k)}$ will be associated with the $c$-th equation in $\Omega^{(k)}$ in (\ref{eq.integMPC_F3}). We further use the expression $i \rightarrow c$ to represent that $U_{(i)}^{(k)}$ is linearly correlated to $Z_{(c)}^{(k)}$ through $A_{in}$, and use $m_{(c)}$ to denote the number of objectives that $Z_{(c)}^{(k)}$ is involved. Then we have:
\begin{equation} \label{eq.F3_dual_1}
\begin{aligned}
    m_{(c)} = \Bigl|\bigcup_{i \rightarrow c} S_{(i)} \Bigr|
\end{aligned}
\end{equation} 

A ``max" operation can be used to handle indicator function $L_+(\bar{Z})$ and the equality constraint $\Omega^{(k)}$ for the global update in algorithm \ref{alg:1}, which can be written as,
\begin{equation} \label{eq.F3_dual}
\begin{aligned}
    Z_{(c)}^{(k)} &= \max
    \begin{Bmatrix*}[l] 0, \displaystyle\frac{1}{m_{(c)}} \displaystyle\sum_{j=1}^{m_{(c)}} (-{\bar A}_{in,j} \bar{U}_{j} - \frac{1}{\rho^j} \gamma^j + \bar{b}_{in,j}) |_{(c)}^{(k)} 
    \end{Bmatrix*} 
\end{aligned}
\end{equation}
where $(\cdot)|_{(c)}^{(k)}$ represent the $(k\cdot N_p + c)$-th element in $(\cdot)$.}

{\relax \textbf{Dual update:}} The update of the Lagrange multiplier in algorithm \ref{alg:1} can be updated as,
\begin{equation} \label{eq.F3_multiplier}
    \gamma_j = \gamma_j + \rho_j ({\bar A}_{in,j} \bar{U}_{j} - \bar{b}_{in,j} + \bar{Z}_j)
\end{equation}

This formulation's biggest advantage is explicitly solving the local optimization for each objective, and the satisfaction of constraints is achieved through the ``max" operation of the virtual center node. Through iterative convergence, the final result can simultaneously meet the requirements of global optimum and constraint satisfaction. 
{\relax  One requirement for this formulation is that each control variable $U_{(i)}$ should have an inequality constraint to be associated with at least one global variable $Z_{(c)}$. This requirement is reasonable in practical engineering because at least there is a min-max constraint for each actuator.}
The discussion about the convergence and the optimal parameter selection can be found in the next section.

{\relax 
\subsection{Convergence and Parameter Selection}

All three formulations strictly follow the ADMM framework, whose convergence has been proven in literature, such as \cite{boyd2011distributed}. The control action $\bar{U}$ generated from these distributed formulations will approximate the global optimal from the integrated MPC, as the residuals $\|r_j^{\langle q \rangle}\|_2^2$ and $\|s_j^{\langle q \rangle}\|_2^2$ converge to 0 through iterations.

$\rho_{j}$ is an important parameter in the proposed multi-objective AMPC because it affects the convergence speed. For the first two formulations, it is difficult to have a deterministic method to determine the value of $\rho_{j}$. So, it is usually selected based on experience in practice. However, for the last formulation, the guidance of choosing a proper $\rho_{j}$ has been proposed in \cite{ghadimi2014optimal} on a similar distributed Quadratic programming (QP) problem by optimizing the upper bound of the converging rate of $\|s_j^{\langle q \rangle}\|_2^2$. 
Similarly, the optimal selection of $\rho_{j}$ for the proposed \textit{Local Explicit} formulation in this paper can be written as,
\begin{equation} \label{eq.opt_ro}
    \rho_{j}^* = \frac{1}{\sqrt{\lambda_1({\bar A}_{in,j} H_j^{-1} {\bar A}_{in,j}^{T})\lambda_n({\bar A}_{in,j} H_j^{-1} {\bar A}_{in,j}^{T})}}
\end{equation}
where $\lambda_1(\cdot) > 0$ and $\lambda_n(\cdot) > 0$ are the minimum non-zero and maximum eigenvalues of the bracketed matrix $(\cdot)$, respectively.}

\section{Examples on Vehicle Control} \label{Vehicle Examples}

Two examples of MACSs are introduced and tested.
As shown in Fig. \ref{exp_vehicle}, the test vehicle has independent torque control on wheels, and its specification is listed in Table \ref{table_param}.

\begin{table}[!ht]
\renewcommand{\arraystretch}{1.3}
\caption{Specification of the Test Vehicle}
\label{table_param}
\centering
\begin{tabular}{c|c|c}
\hline\hline
Parameter & Description & Value \\
\hline
$m$ & Vehicle mass & $2272$ kg \\
$I_z$ & Vehicle inertia & $4600~\text{kg}\cdot{\text{m}}^2$ \\
$I_w$ & Vehicle track width & $1.6$ m \\
$a$ & Front distance from axles to CG & $1.42$ m \\
$b$ & Rear distance from axles to CG & $1.43$ m \\
$L$ & Wheel base & $2.85$ m \\
$R_w$ & Effective radius of wheels & $0.351$ m \\
$K_{us}$ & Desired under-steering gradient & $0.005$ $\text{s}^2/\text{m}$\\
$T_w$ & Wheel track & $1.66$ m\\
\hline\hline
\end{tabular}
\end{table}

\subsection{Cruise Control} \label{exp_intro_1}

The longitudinal and lateral objectives should be met simultaneously in cruise control. The former is a task of longitudinal speed tracking ($j=$ LST), while the latter is a stability control of yaw rate tracking ($j=$ YRT). {\relax In this example, we have four agents: the rear-wheel-drive (RWD) driving torque ($i=$ DTR) agent generates the driving torque on the rear axle; the rear torque vectoring ($i=$ RTV) agent generates the opposite torques on the left and right wheels of the rear axle; the front differential braking ($i=$ FDB) agent generates individual braking torques on the front axle; the driver steering ($i=$ STR) agent generates the front steering angle. In cruise control systems, the STR agent is uncontrollable for electronic controllers since the driver controls the steering, and its contribution is treated as an external disturbance in $W_d$. The configuration is shown in Fig. \ref{fig_example_1}, where the agents' contributions are obtained by analyzing the following system equations.
In this example, The control actions from each agent are considered as, 
\begin{equation} \label{eq.CC_system_1}
\renewcommand{\arraystretch}{1.2}
\begin{array}{l}
\text{DTR:} ~ U_{\text{(DTR)}} = T_{\text{DTR}} \in [-2000, 2000]\\
\text{RTV:} ~ U_{\text{(RTV)}} = T_{\text{RTV}} \in [-500, 500]\\ 
\text{FDB:} ~ U_{\text{(FDB)}} = {[T_{\text{FDB-L}}, T_{\text{FDB-R}}]}^T \in [-1000, 0]\\
\text{STR:} ~ U_{\text{(STR)}} = {[F_{y-\text{FL}}, F_{y-\text{FR}}, F_{y-\text{RL}}, F_{y-\text{RR}}]}^T \\
\end{array}
\end{equation}
where, $T_{\text{DTR}}$ is the total drive torque. $T_{\text{RTV}}$ is the differential torque applied on the right-rear wheel, while the opposite is applied on the left. 
$T_{\text{FDB-L}}$ and $T_{\text{FDB-R}}$ are the independent braking torque applied on the right-front and left-front wheels, respectively. $F_y$ are the estimated lateral tire forces.

\begin{figure}[!t]
\centering
\includegraphics[width=3.1in]{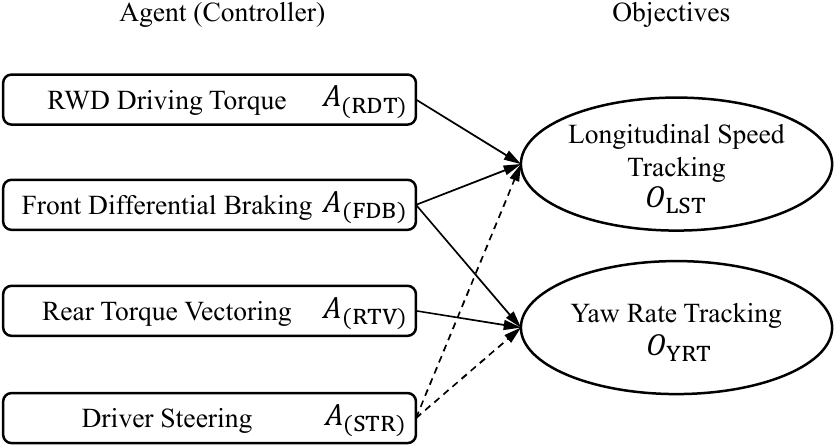}
\caption{{\relax Example of the MACS for cruise control. The two objectives share the FDB agent. The STR agent is the external input on both objectives, so its contributions are drawn as dashed arrows. Thus, $S_{\text{YRT}} = \{\text{RTV}, \text{FDB}\}$, $S_{\text{LST}} = \{\text{DTR}, \text{FDB}\}$. For agents: $S_{(\text{DTR})} = \{\text{LST}\}$, $S_{(\text{FDB})} = \{\text{YRT}, \text{LST}\}$, $S_{(\text{RTV})} = \{\text{YRT}\}$.}}
\setlength{\tabcolsep}{3pt}
\label{fig_example_1}
\end{figure}

State $X$ is considered as the longitudinal speed ($v_x$), lateral speed ($v_y$), and yaw rate ($\dot\phi$). Then, for the objectives, we have:
\begin{equation} \label{eq.CC_objective}
\renewcommand{\arraystretch}{1.3}
\begin{array}{c}
X_{ref} = [v_{x,ref}, 0, \dot\phi_{ref}], \\
\dot\phi_{ref} = \text{sign}(\delta_f) \min \left( |\frac{v_x}{L+K_{us}v_x^2} \delta_f |, \frac{\mu g}{v_x} \right), \\
C_{\text{YRT}} = {[1, 0, 0]}, ~ C_{\text{LST}} = {[0, 0, 1]}
\end{array}
\end{equation}
where $\delta_f$ is the input of the driver's steering angle. $\mu$ is the road friction. $g$ is the gravity acceleration. $v_{x, ref}$ and $\dot\phi_{ref}$ are the target longitudinal speed from the upper-level speed planner and the target yaw rate calculated from the single-track vehicle model and limited by the road friction, respectively.

The system matrices in (\ref{eq_linear}) for cruise control could be written as,
\begin{equation} \label{eq.CC_system}
\renewcommand{\arraystretch}{1.3}
\begin{array}{l}
\setlength{\tabcolsep}{0.5pt}   
A_d=\left[ \begin{tabular}{ c c c } $1$ & $\dot\phi \Delta t$ & $v_y\Delta t$ \\
                        $-\dot\phi \Delta t$ & $1$ & $-v_x \Delta t$ \\
                        $0$ & $0$ & $1$ \end{tabular} \right], \\
B_{d(\text{DTR})} = {[{\Delta t}/(mR_w), 0, 0]}^T,\\
B_{d(\text{RTV})} = {[0, 0, \Delta t T_w/(I_zR_w)]}^T,\\ 
B_{d(\text{FDB})} = {\Delta t} \cdot \\ 
\setlength{\tabcolsep}{1.3pt} 
\left[ \begin{tabular}{c c } 
                        $\cos{\delta_{\text{FL}}}/(mR_w)$ & 
                        $\cos{\delta_{\text{FR}}}/(mR_w)$ \\
                        $\sin{\delta_{\text{FL}}}/(mR_w)$ & 
                        $\sin{\delta_{\text{FR}}}/(mR_w)$ \\
                        $\displaystyle\frac{2a\sin{\delta_{\text{FL}}}-T_w\cos{\delta_{\text{FL}}}}{2I_zR_w}$ & 
                        $\displaystyle\frac{2a\sin{\delta_{\text{FR}}}+
                        T_w\cos{\delta_{\text{FR}}}}{2I_zR_w}$ 
                        \end{tabular} \right], \\
W_d = B_{d,\text{STR}}U_{\text{STR}}, \\
B_{d(\text{STR})} = {\Delta t} \cdot\\ 
\setlength{\tabcolsep}{2.0pt} 
\left[ \begin{tabular}{c c c c} 
                        $-\frac{1}{m}\sin{\delta_{\text{FL}}}$ & $-\frac{1}{m}\sin{\delta_{\text{FR}}}$ & $0$ & $0$ \\
                        $\frac{1}{m}\cos{\delta_{\text{FL}}}$ & $\frac{1}{m}\cos{\delta_{\text{FR}}}$ & $\frac{1}{m}$ & $\frac{1}{m}$ \\
                        $\renewcommand{\arraystretch}{1.0} 
                        \setlength{\tabcolsep}{0.5pt}
                        \Bigl(\begin{array}{l}
                             \frac{a}{I_z}\cos{\delta_{\text{FL}}} \\
                             +\frac{T_w}{2I_z}\sin{\delta_{\text{FL}}}
                        \end{array} \Bigr)$ & 
                        $\renewcommand{\arraystretch}{1.0} 
                        \setlength{\tabcolsep}{0.5pt}
                        \Bigl(\begin{array}{l}
                             \frac{a}{I_z}\cos{\delta_{\text{FR}}} \\
                             -\frac{T_w}{2I_z}\sin{\delta_{\text{FR}}}
                        \end{array} \Bigr)$ & 
                        $-\frac{b}{I_z}$ & $-\frac{b}{I_z}$
                        \end{tabular} \right].
\end{array}
\end{equation}
where $\Delta t$ is the sample time. $\delta_{\text{FR}}$ and $\delta_{\text{LR}}$ are right and left wheel steering angles at the front axle, respectively.
}

\subsection{\relax Holistic Vehicle Stability Control} \label{exp_intro_2}

{\relax Holistic vehicle stability control is another challenging task that requires coordinating multiple controllers to meet various objectives. In this example, the objective of vehicle stability control (VSC) tracks the desired yaw rate, and the wheel stability control (WSC) tracks the desired wheel speed for maximum tire capability.}
{\relax This example has one more agent than the \textit{Cruise Control}: the traction control rear ($i=$ TCR) agent will reduce the driving torque on the rear axle if the rear wheels start to slip. 
The control actions generated from agent TCR are considered as $\text{TCR:} ~ U_{\text{(TCR)}} = T_{\text{TCR}} \in [0, T_{\text{DTR}}]$, 
where, $T_{\text{TCR}}$ is reduction in driving torque.
In this example, the STR and DTR agents are uncontrollable, and their contributions are treated as external disturbances in objective VSC and WSC. 
The configuration is shown in Fig. \ref{fig_example_2}. 

\begin{figure}[!t]
\centering
\includegraphics[width=3.1in]{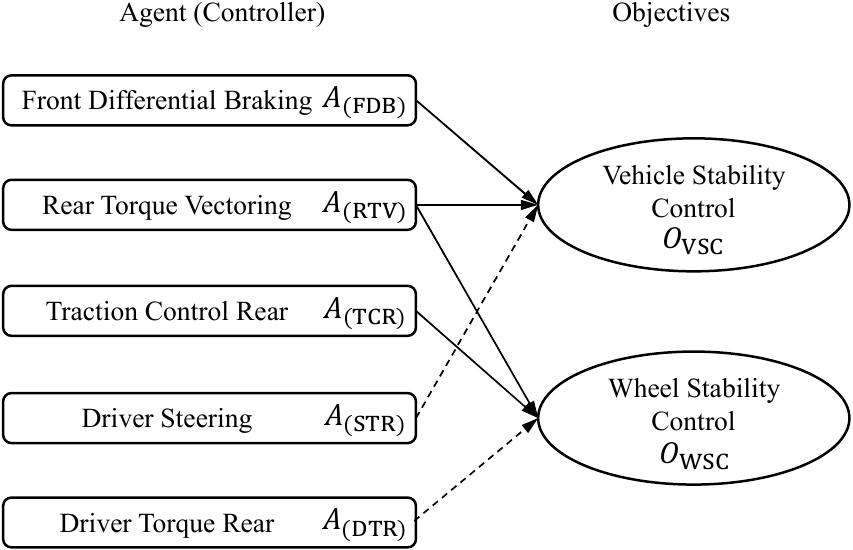}
\caption{{\relax Example of the MACS for holistic vehicle stability control. The two objectives share the RTV agent. The STR and DTR agents are the external input on each objective, respectively, so their contributions are drawn as dashed arrows. Thus, $S_{\text{VSC}} = \{\text{RTV}, \text{FDB}\}$, $S_{\text{WSC}} = \{\text{TCR}, \text{RTV}\}$. For agents: $S_{(\text{FBD})} = \{\text{VSC}\}$, $S_{(\text{RTV})} = \{\text{VSC}, \text{WSC}\}$, $S_{(\text{TCR})} = \{\text{WSC}\}$.}}
\setlength{\tabcolsep}{3pt}
\label{fig_example_2}
\end{figure}

State $X$ is considered as the lateral speed ($v_y$), yaw rate ($\dot\phi$), and speeds of rear wheels ($\omega_{\text{RL}}$, $\omega_{\text{RR}}$). Then,
the system matrices in (\ref{eq_linear}) for holistic vehicle control could be written as,
{\relax 
\begin{equation} \label{eq.VSC_system}
\renewcommand{\arraystretch}{1.2}
\begin{array}{l}
\setlength{\tabcolsep}{1.5pt}   
A_d=\left[ \begin{tabular}{ c c c c } $1$ & $-v_x \Delta t$ & $0$ & $0$ \\
                        $0$ & $1$ & $0$ & $0$ \\
                        $0$ & $0$ & $1$ & $0$ \\
                        $0$ & $0$ & $0$ & $1$ \end{tabular} \right],\\
B_{d(\text{TCR})} = {[0, 0, -{\Delta t}/(2I_w), -{\Delta t}/(2I_w)]}^T, \\
B_{d(\text{RTV})} = {[0, {\Delta t}T_z/R_w, -{\Delta t}/I_w, {\Delta t}/I_w]}^T, \\ 
B_{d(\text{FDB})} = {\Delta t} \cdot\\ 
\renewcommand{\arraystretch}{1.0}
\setlength{\tabcolsep}{1.0pt} 
\left[ \begin{tabular}{c c } 
                        $\sin{\delta_{\text{FL}}}/{mR_w}$ & 
                        $\sin{\delta_{\text{FR}}}/(mR_w)$ \\
                        $\displaystyle\frac{2a\sin{\delta_{\text{FL}}}-T_w\cos{\delta_{\text{FL}}}}{2I_zR_w}$ & $\displaystyle\frac{2a\sin{\delta_{\text{FR}}}+
                        T_w\cos{\delta_{\text{FR}}}}{2I_zR_w}$ \\
                        $0$ & $0$ \\
                        $0$ & $0$ \\
                        \end{tabular} \right], \\
W_d^k = B_{d,\text{DTR}}U_{\text{DTR}} + B_{d,\text{STR}}U_{\text{STR}}, \\
\end{array}
\end{equation}
and,
\begin{equation} \label{eq.VSC_system_1}
\renewcommand{\arraystretch}{1.2}
\begin{array}{l}
B_{d(\text{DTR})} = {[0, 0, {\Delta t}/(2I_w), {\Delta t}/(2I_w)]}^T,\\ 
B_{d(\text{STR})} = {\Delta t} \cdot\\ 
\setlength{\tabcolsep}{2.0pt} 
\left[ \begin{tabular}{c c c c}                         
                        $\frac{1}{m}\cos{\delta_{\text{FL}}}$ & $\frac{1}{m}\cos{\delta_{\text{FR}}}$ & $\frac{1}{m}$ & $\frac{1}{m}$ \\
                        $\renewcommand{\arraystretch}{1.0} 
                        \setlength{\tabcolsep}{0.5pt}
                        \Bigl(\begin{array}{l}
                             \frac{a}{I_z}\cos{\delta_{\text{FL}}} \\
                             +\frac{T_w}{2I_z}\sin{\delta_{\text{FL}}}
                        \end{array} \Bigr)$ & 
                        $\renewcommand{\arraystretch}{1.0} 
                        \setlength{\tabcolsep}{0.5pt}
                        \Bigl(\begin{array}{l}
                             \frac{a}{I_z}\cos{\delta_{\text{FR}}} \\
                             -\frac{T_w}{2I_z}\sin{\delta_{\text{FR}}}
                        \end{array} \Bigr)$ & 
                        $-\frac{b}{I_z}$ & $-\frac{b}{I_z}$ \\
                        $0$ & $0$ & $0$ & $0$ \\
                        $0$ & $0$ & $0$ & $0$ 
                        \end{tabular} \right].
\end{array}
\end{equation}
}

For the objectives, we have:
\begin{equation} \label{eq.VSC_objective}
\renewcommand{\arraystretch}{1.0}
\begin{array}{c}
X_{ref} = [0, \dot\phi_{ref}, \omega_{ref}, \omega_{ref}],\\
\omega_{ref} = v_x(1+S_{x,des}) \cdot [1,1]^T \\
C_{\text{VSC}} = {[0, 1, 0, 0]}, 
\setlength{\tabcolsep}{1.5pt} 
~ C_{\text{WSC}} = \left[ \begin{tabular}{ c c c c } 
                        $0$, & $0$, & $1$, & $0$ \\
                        $0$, & $0$, & $0$, & $1$ \end{tabular} \right],\\
\end{array}
\end{equation}
where $\delta_f$ and $v_x$ are inputs of the driver's steering angle at front wheels and vehicle speed, respectively. The target yaw rate $\dot\phi_{ref}$ is the same as in (\ref{eq.CC_objective}). $S_{x,des}$ is the desired longitudinal slip ratio, which usually is $15\%$.

}

\section{Experimental Studies} \label{Examples}



{\relax All simulations and experiments run in Matlab in real time. 
All real-world experiments are implemented on a real vehicle with four independent motors that could simulate various controllers, as shown in Fig. \ref{exp_vehicle}. Our tests were conducted on a dedicated test track that provides different road frictions and enough space for aggressive driving. The controllers were running in real-time at 20Hz on a laptop with an i7-12700H CPU and 16 GB RAM. The controller ran in the Matlab Real-time Desktop model and sent the control command to motor drivers through the CAN bus at every time step. For simulation, A high-fidelity vehicle model was used in CarSim.} 

Two comparisons were conducted for each experiment: (1) Comparison between partially disabling one objective and enabling all the objectives to verify the effectiveness of the proposed multi-objective AMPC scheme. (2) Comparison between the multi-objective AMPC and integrated MPC to demonstrate convergence and improved computational efficiency. 
The proposed three formulations are compared in the \textit{Cruise Control} example using simulations. 
{\relax In experimental comparisons to the integrated MPC, only the \textit{Local Explicit} formulation is used since it is more efficient than the other two formulations. }
The time consumption in all comparisons only counts the core processes of solving MPCs, excluding matrix initialization and post-processing.

\subsection{Example 1: Cruise Control} \label{CC}

{\relax 
Aggressive lane changes were tested on high-friction road surfaces using Section \ref{exp_intro_1} configuration. A double line change (DLC) during acceleration was conducted following a single line change (SLC) during deceleration. 

\textbf{Simulations: Comparison of three formulations}}
To evaluate the convergence performance, we first define the suboptimality as the absolute error between the control actions between the multi-objective AMPC and the integrated MPC:
\begin{equation} \label{eq.suboptimality}
    e_{sub} = |U_{distributed}^{(0)} - U_{integrated}^{(0)}|
\end{equation}
where $U_{distributed}^{(0)}$ and $U_{integrated}^{(0)}$ are the output control actions from the multi-objective AMPC and the integrated MPC, respectively. 
We also define $r_{time}$ as the time consumption ratio of the distributed form and the centralized form:
\begin{equation} \label{eq.t_ratio}
    r_{time} = \frac{t_{distributed}}{t_{integrated}}
\end{equation}
where $t_{distributed}$ and $t_{integrated}$ are the time consumption using the distributed and integrated formulations, respectively. 

\begin{figure}[!t]
\centering
\includegraphics[width=3.5in]{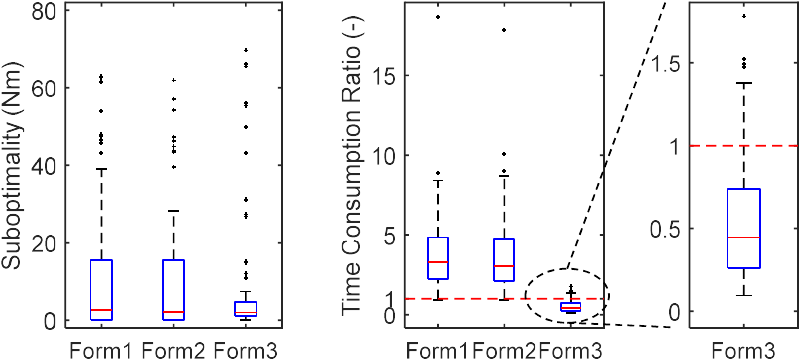}
\caption{\relax Comparison results between three distributed formulations proposed in Section III: Form 1 (\textit{Fully Distributed}), Form 2 (\textit{Virtual Center Node}, and Form 3 (\textit{Local Explicit}). The suboptimality is the absolute error of the torque control actions between the multi-objective AMPC and the integrated MPC. The formulation of multi-objective AMPC is said to be more computationally efficient than the integrated MPC if the time consumption ratio is less than 1.} 
\setlength{\tabcolsep}{3pt}
\label{fig_compare}
\end{figure}

Using different formulations, figure \ref{fig_compare} compares the suboptimality and time consumption ratios. 
{\relax The parameter $\rho_j$ is manually selected for the first two formulations, \textit{Fully Distributed} and \textit{Virtual Center Node}, to achieve the best efficiency, and it is calculated using (\ref{eq.opt_ro}) for the \textit{Local Explicit} formulation.}
{\relax As shown in the results, all three formulations can approximate the global optimum since the majority of the torque suboptimalities are less than 20 Nm (5 Nm for \textit{Local Explicit}), which is completely acceptable and could even be ignored in vehicle control.} 
Besides, the \textit{Local Explicit} formulation achieved the best computational efficiency and even ran faster than the integrated formulation with $r_{time} < 1$.

\begin{figure}[!t]
\centering
\includegraphics[width=3.5in]{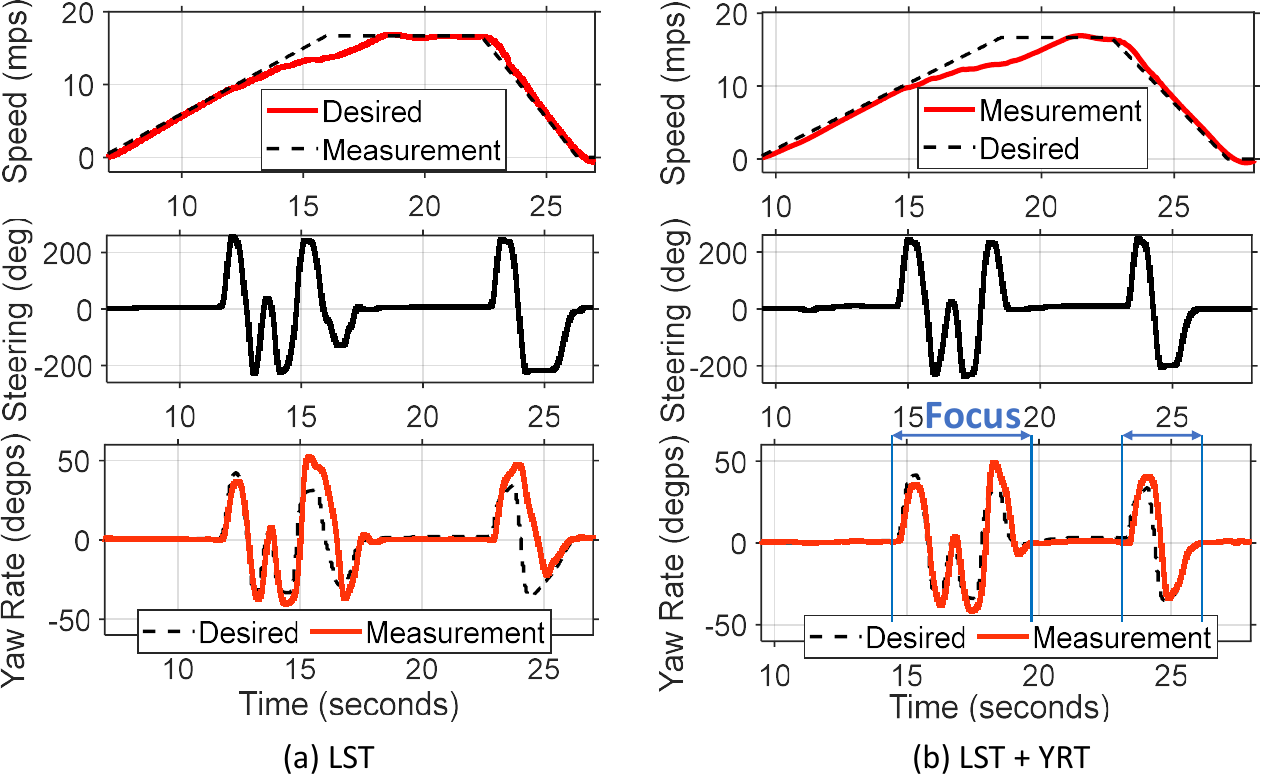}
\caption{Comparison of the vehicle response for the aggressive lane change in the example of cruise control. The over-steering was significantly suppressed when the objective of YRT was enabled. The yaw stability was improved.}
\setlength{\tabcolsep}{3pt}
\label{fig_CC_vehicle}
\end{figure}

\textbf{Experiments: Aggressive lane change} {\relax  Two comparative tests were conducted: only the objective of LST was enabled in the first experiment; both objectives were enabled in the second experiment.} The \textit{Local Explicit} formulation with the optimal parameters was used in the second experiment for the multi-objective MPC. Those two tests had the same desired speed profile and similar steering maneuvers. The results of the vehicle response are shown in Fig. \ref{fig_CC_vehicle}. The acceleration and deceleration are smooth when tracking the target speed in both tests. The error between the target and measured yaw rates is significantly reduced when the objective of YRT is enabled, indicating the improved vehicle's yaw stability. 

\begin{figure}[!t]
\centering
\includegraphics[width=3.5in]{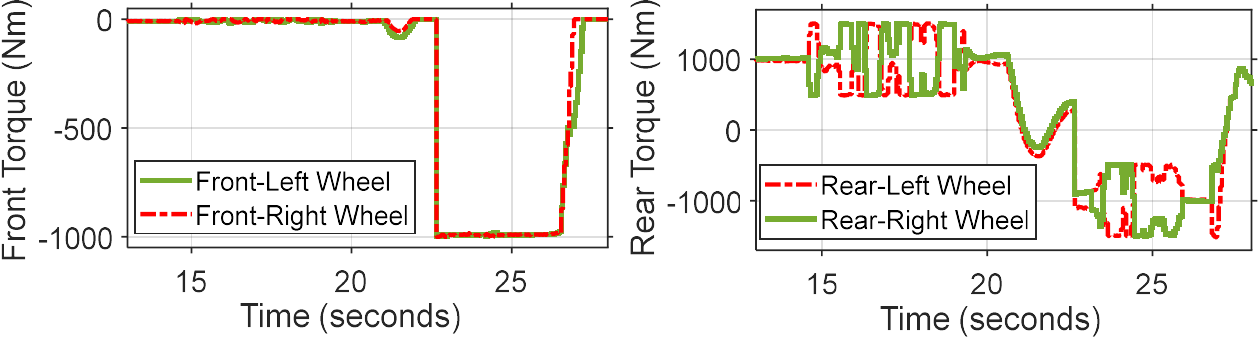}
\caption{\relax Wheel torque commands from the multi-objective AMPC using the \textit{Local Explicit} formulation with optimal parameters. All control agents have cooperated: DTR generated the driving and braking torque on the rear wheels; RTV generated differential torque on the rear wheels; FDB generated independent braking torques on the front wheels.}
\setlength{\tabcolsep}{3pt}
\label{fig_CC_torque}
\end{figure}

The results of torque commands on each wheel are shown in Fig. \ref{fig_CC_torque}, proving that the proposed multi-objective AMPC can successfully collaborate with multiple controllers to achieve multiple objectives simultaneously.

\begin{figure}[!t]
\centering
\includegraphics[width=3.5in]{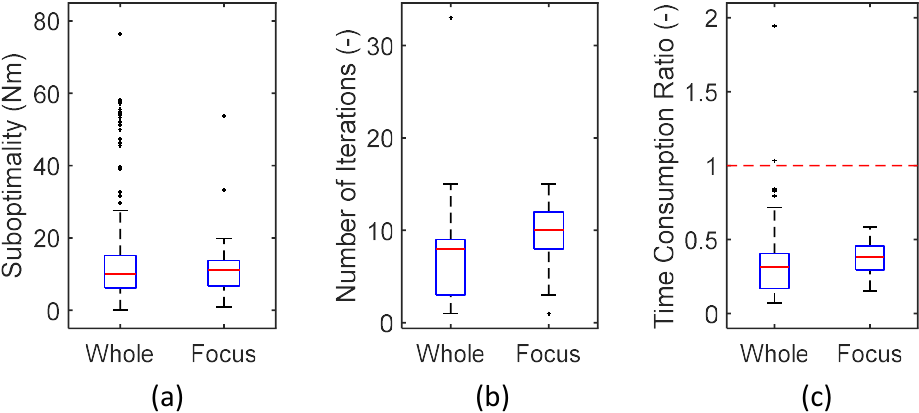}
\caption{Results of the convergence and computational efficiency: (a) torque suboptimality (b) number of iterations (c) time consumption ratio. ``Whole" represents the whole test process, while ``Focus" represents the specific period marked in Fig. \ref{fig_CC_vehicle} where both objectives require significant control.}
\setlength{\tabcolsep}{3pt}
\label{fig_CC_time}
\end{figure}

The suboptimality, iterations, and time consumption ratio results are shown in Fig. \ref{fig_CC_time}. The majority of the torque suboptimality was always less than 20 Nm. The number of iterations required less than 15. The time consumption ratio was significantly less than 1. Besides, it is easy to notice that a few more iterations are needed during the ``focus" period to achieve convergence. Accordingly, the time consumption ratio is slightly higher but less than 1. 
{\relax The results demonstrate that even for such a small system, the proposed multi-objective AMPC in \textit{Local Explicit} formulation with the optimal parameters is computationally efficient and outperforms the integrated MPC while always approximating to the global optimum.}

\subsection{Example 2: Holistic Vehicle Stability Control} \label{SC}

{\relax 
Two maneuvers were experimented with for the holistic vehicle stability control using Section \ref{exp_intro_2} configuration. Two tests were implemented for each experiment: only the driver controlled the vehicle in the first test, while all the electric controllers were enabled in the second test. The multi-objective AMPC in \textit{Local Explicit} formulation with the optimal parameters was used in the second test.}

\begin{figure}[!t]
\centering
\includegraphics[width=3.5in]{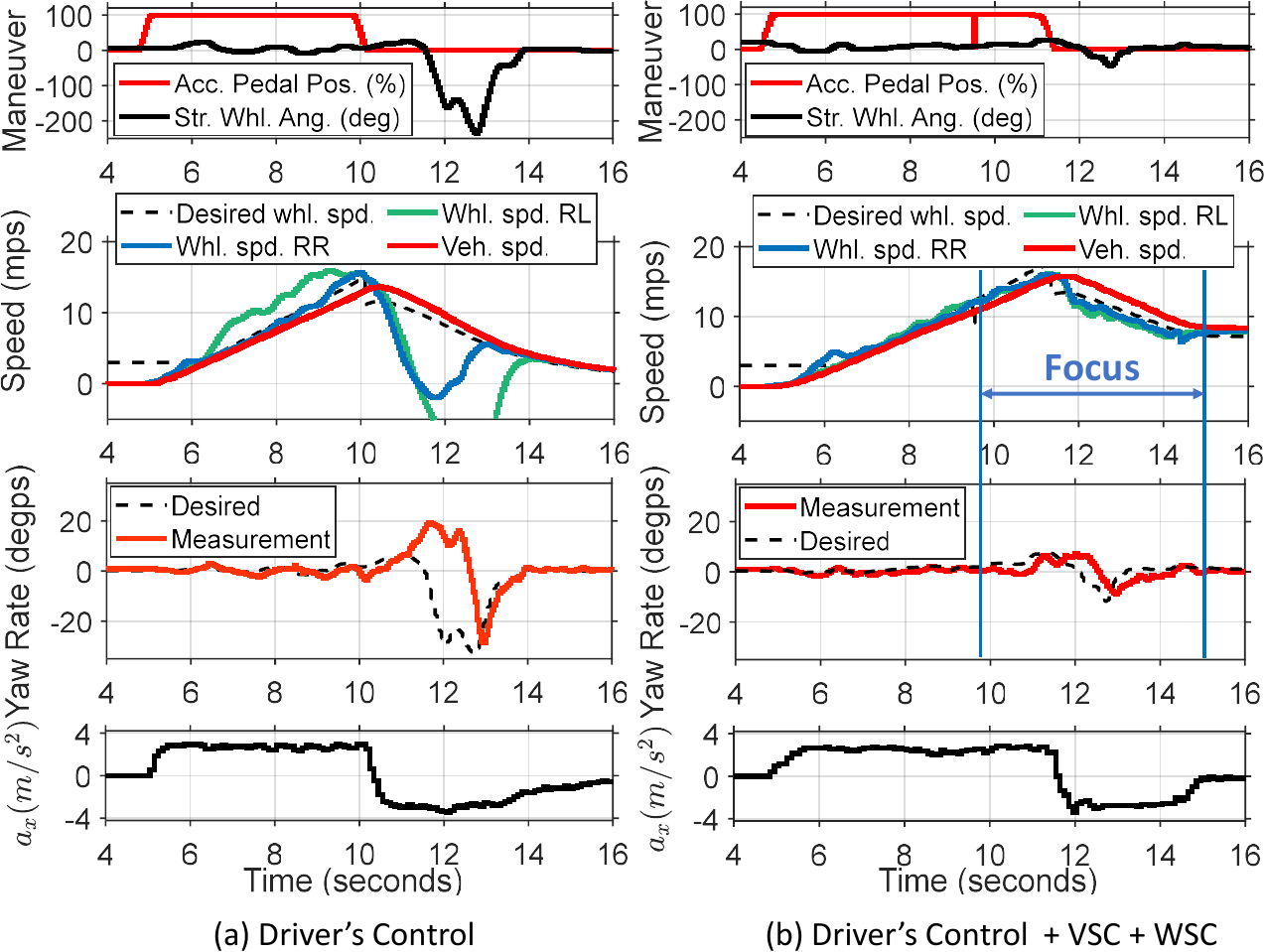}
\caption{Comparison of the vehicle response for straight launch and braking in the example of holistic stability control. The over-steering was significantly suppressed, and no obvious wheel slip occurred during the second test. The holistic stability was improved when both VSC and WSC were enabled.}
\setlength{\tabcolsep}{3pt}
\label{fig_VSC_straight_vehicle}
\end{figure}

\textbf{Experiment 1: Straight launch and brake} {\relax  The vehicle was straight launched and then braked on a low-friction sandy road surface.} The compared results of vehicle response are shown in Fig. \ref{fig_VSC_straight_vehicle}. 
Severe wheel slip and yaw rate error could be observed in the first test due to low road friction. The driver's aggressive steering also shows that a large effort was required to keep the vehicle straight. 
However, when both stability control objectives were enabled in the second test, wheel speeds were controlled at desired values during the launching and braking processes. The yaw rate error was also reduced. The overall stability of the vehicle has been significantly improved. Besides, vehicle stability control does not lose the longitudinal acceleration or deceleration.

\begin{figure}[!t]
\centering
\includegraphics[width=3.5in]{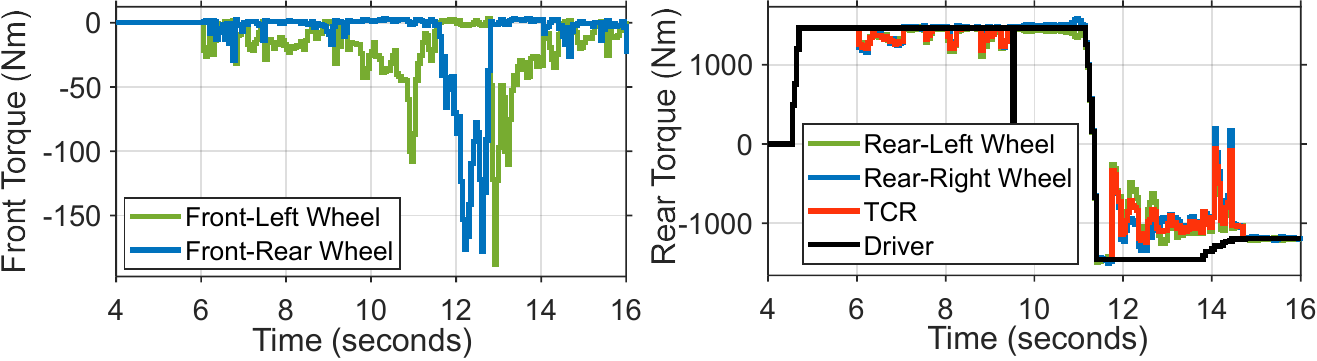}
\caption{Torque commands from the multi-objective AMPC using the \textit{Local Explicit} formulation with optimal parameters. All control agents cooperated: FDB generated independent braking torques on the front wheels; the drive's torque was first reduced by TCR and then adjusted by RTV on the rear wheels.}
\setlength{\tabcolsep}{3pt}
\label{fig_VSC_straight_compare_torque}
\end{figure}

{\relax 
The torques on four wheels in the second test are shown in Fig. \ref{fig_VSC_straight_compare_torque}. The FDB agent generated independent braking torques on the front wheels. The TCR agent first reduced the drive's torque to depress the wheel slip and then adjusted by the RTV agent for both objectives.}

\begin{figure}[!t]
\centering
\includegraphics[width=3.5in]{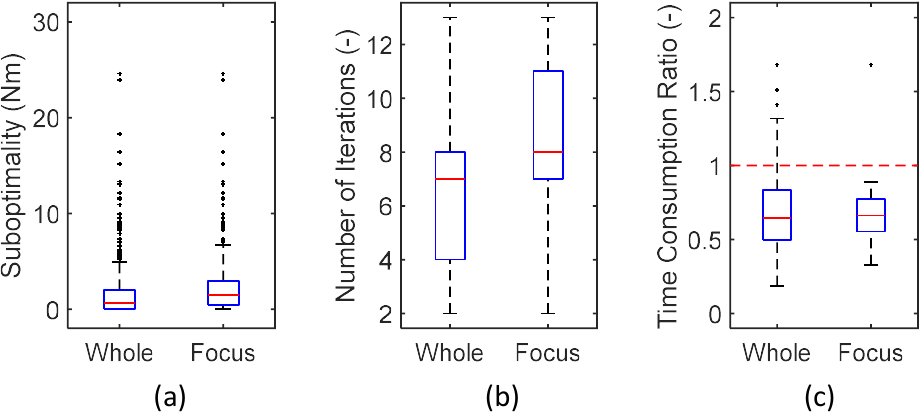}
\caption{Results of the convergence and computational efficiency: (a) torque suboptimality (b) number of iterations (c) time consumption ratio. ``Whole" represents the whole test process, while ``Focus" represents the specific period marked in Fig. \ref{fig_VSC_straight_vehicle} where both objectives require significant control.}
\setlength{\tabcolsep}{3pt}
\label{fig_VSC_straight_time}
\end{figure}

The suboptimality, iterations, and time consumption ratio results are shown in Fig. \ref{fig_VSC_straight_time}. The majority of the torque suboptimality was less than 5 Nm. The number of iterations required was less than 15. The time consumption ratio was less than 1. A few more iterations are needed to achieve convergence during the ``focus" period, so the time consumption ratio is slightly higher but still less than 1. 

\begin{figure}[!t]
\centering
\includegraphics[width=3.5in]{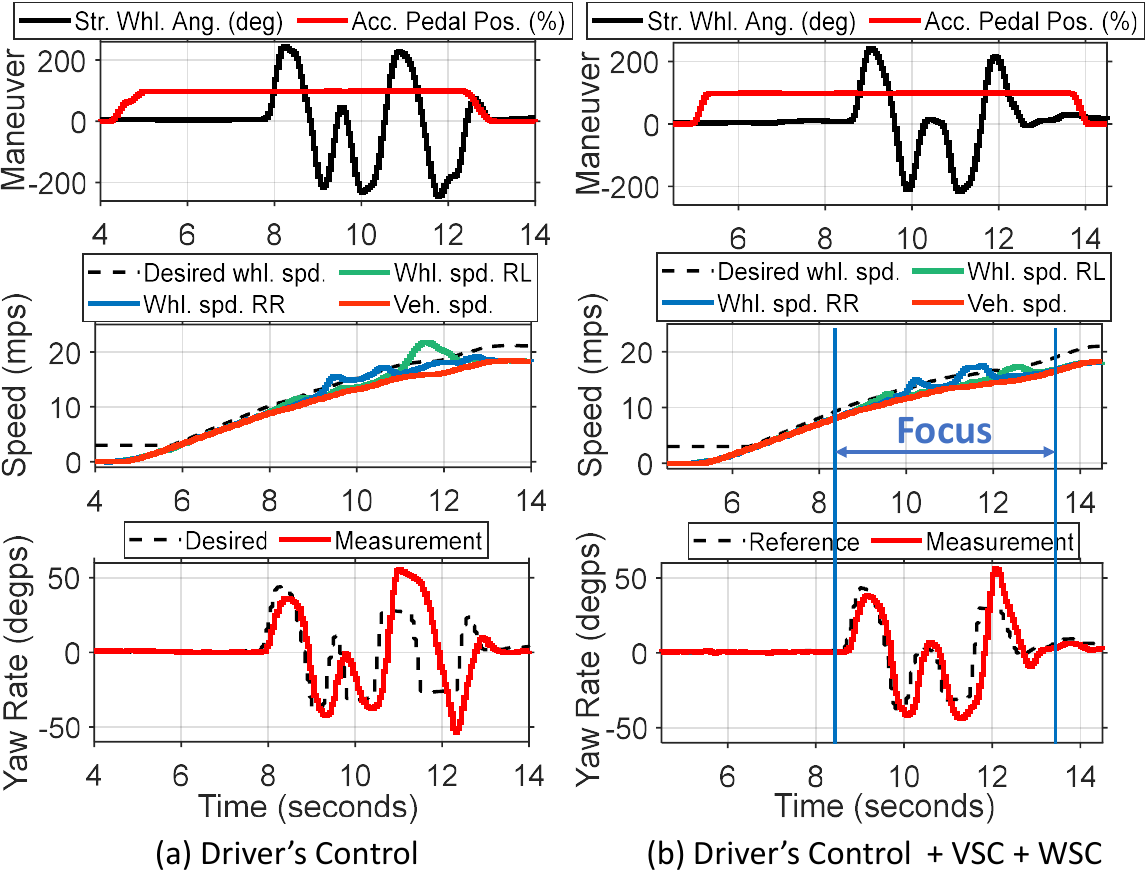}
\caption{Comparison of the vehicle response for DLC acceleration in the example of holistic stability control. The over-steering was significantly suppressed, and no obvious wheel slip occurred during the second test. The holistic stability was improved when both VSC and WSC were enabled.}
\setlength{\tabcolsep}{3pt}
\label{fig_VSC_DLC_vehicle}
\end{figure}

\textbf{Experiment 2: Double lane change when acceleration} {\relax A DLC was conducted during the acceleration on high-friction road surfaces.
The compared vehicle performance is shown in Fig. \ref{fig_VSC_DLC_vehicle}. 
Significant longitudinal wheel slip and over-steering happened in the first test, where the driver needed aggressive counter-steering to stabilize the vehicle. However, when both objectives were enabled in the second test, yaw rate error was significantly reduced, with no excessive slip.}

\begin{figure}[!t]
\centering
\includegraphics[width=3.5in]{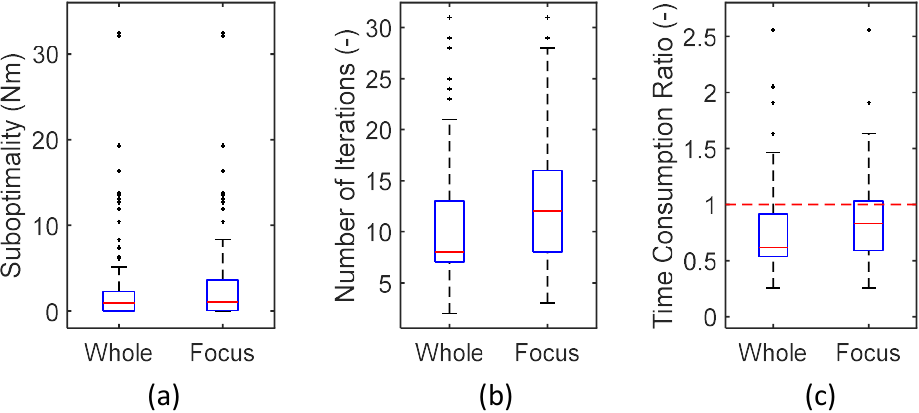}
\caption{Results of the convergence and computational efficiency: (a) torque suboptimality (b) number of iterations (c) time consumption ratio. ``Whole" represents the whole test process, while ``Focus" represents the specific period marked in Fig. \ref{fig_VSC_DLC_vehicle} where both objectives require significant control.}
\setlength{\tabcolsep}{3pt}
\label{fig_VSC_DLC_time}
\end{figure}

The suboptimality, iterations, and time consumption ratio results are shown in Fig. \ref{fig_VSC_DLC_time}. The majority of the torque suboptimality was always less than 5 Nm. The number of iterations required was less than 25. The time consumption ratio was less than 1.

\section{Conclusion} \label{Conclusion}

This paper introduces a novel multi-objective AMPC scheme for ``plug-and-play" vehicle control. Three formulations tailored from ADMM were systematically developed and applied to vehicle control systems. 
{\relax These formulations approximate the global optimum while managing control regularization and linear inequality constraints.} 
The proposed formulations were evaluated through simulations, demonstrating that the \textit{Local Explicit} formulation offers the highest computational efficiency and even outperforms the integrated MPC. Additionally, the method was experimentally validated using two typical MACS examples in vehicle control. The overall vehicle performance was improved by achieving multiple objectives simultaneously. 
{\relax The results confirmed the approximate global convergence and superior computational efficiency of the \textit{Local Explicit} formulation.}
In conclusion, the multi-objective AMPC introduced in this paper offers a practical and versatile distributed solution for complex vehicle control systems:
\begin{itemize}
    \item[a)] It enhances design flexibility by decoupling the system based on control objectives.
    \item[b)] {\relax It approximates global optimization while meeting the standard requirements of real-world control systems.}
    \item[c)] It also has the potential to provide superior real-time computational efficiency compared to integrated MPC.
\end{itemize}

The proposed scheme has the potential to be extended beyond single-vehicle MACSs, serving as a general distributed control approach for systems that are physically distributed. This includes applications like interaction-aware planning, multi-vehicle collision avoidance, and fleet management. Future works will focus on expanding the proposed method to these areas, particularly for MACSs with higher degrees of non-linearity.

{\relax Finally, extending ADMM beyond two groups is non-trivial and still an ongoing research topic. For example, some recent studies have demonstrated that the convergence of ADMM can be guaranteed only under additional structural assumptions \cite{chen2016direct}\cite{hong2017linear}. The simulation and real vehicle test examples in this paper only involve the decomposition of two groups, which may be insufficient given the development of intelligent vehicle features. Thus, in the future, exploring the AMPC method that could be applied to more than two groups would be a valuable research direction.}


\section*{Acknowledgment}

The authors would like to acknowledge the financial support of the Natural Sciences and Engineering Research Council (NSERC) of Canada in this work.

\bibliographystyle{bibtex/IEEEtran}
\bibliography{bibtex/IEEEabrv,Paper/ref}

@article{mazzilli2021integrated,
  title={Integrated chassis control: Classification, analysis and future trends},
  author={Mazzilli, Victor and De Pinto, Stefano and Pascali, Leonardo and Contrino, Michele and Bottiglione, Francesco and Mantriota, Giacomo and Gruber, Patrick and Sorniotti, Aldo},
  journal={Annu. Rev. Control},
  volume={51},
  pages={172--205},
  year={2021},
  publisher={Elsevier}
}

@inproceedings{elliott2008model,
  title={Model-based predictive control of a multi-evaporator vapor compression cooling cycle},
  author={Elliott, Matthew S and Rasmussen, Bryan P},
  booktitle={Proc. Amer. Control Conf.},
  pages={1463--1468},
  year={2008},
  organization={IEEE}
}

@article{liu2010sequential,
  title={Sequential and iterative architectures for distributed model predictive control of nonlinear process systems},
  author={Liu, Jinfeng and Chen, Xianzhong and Mu{\~n}oz de la Pe{\~n}a, David and Christofides, Panagiotis D},
  journal={AIChE J.},
  volume={56},
  number={8},
  pages={2137--2149},
  year={2010},
  publisher={Wiley Online Library}
}

@article{maestre2011distributed,
  title={Distributed model predictive control based on agent negotiation},
  author={Maestre, JM and De La Pena, D Munoz and Camacho, EF and Alamo, T},
  journal={J. Process Control},
  volume={21},
  number={5},
  pages={685--697},
  year={2011},
  publisher={Elsevier}
}

@article{tang2021agent,
  title={Agent-based model predictive controller (AMPC) for flexible and efficient vehicular control},
  author={Tang, Chen and Khajepour, Amir},
  journal=IEEE_J_VT,
  volume={70},
  number={10},
  pages={9877--9885},
  year={2021},
  publisher={IEEE}
}

@article{tang2020wheel,
  title={Wheel modules with distributed controllers: A multi-agent approach to vehicular control},
  author={Tang, Chen and Khajepour, Amir},
  journal=IEEE_J_VT,
  volume={69},
  number={10},
  pages={10879--10888},
  year={2020},
  publisher={IEEE}
}

@article{boyd2011distributed,
  title={Distributed optimization and statistical learning via the alternating direction method of multipliers},
  author={Boyd, Stephen and Parikh, Neal and Chu, Eric and Peleato, Borja and Eckstein, Jonathan and others},
  journal={Found. Trends Mach. learn.},
  volume={3},
  number={1},
  pages={1--122},
  year={2011},
  publisher={Now Publishers, Inc.}
}

@article{zhou2023distributed,
  title={Distributed model predictive control methods for intermodal transport cooperative planning based on ADMM},
  author={Zhou, Qicai and Huang, Yuankai and Xiong, Xiaolei and Zhao, Jiong},
  journal={IET Intell. Transp. Syst.},
  volume={17},
  number={1},
  pages={102--118},
  year={2023},
  publisher={Wiley Online Library}
}

@inproceedings{van2016online,
  title={Online distributed motion planning for multi-vehicle systems},
  author={Van Parys, Ruben and Pipeleers, Goele},
  booktitle={2016 Eur. Control. Conf.},
  pages={1580--1585},
  year={2016},
  organization={IEEE}
}

@article{shi2021distributed,
  title={Distributed model predictive control for joint coordination of demand response and optimal power flow with renewables in smart grid},
  author={Shi, Ye and Tuan, Hoang Duong and Savkin, Andrey V and Lin, Chin-Teng and Zhu, Jian Guo and Poor, H Vincent},
  journal={Appl. Energy},
  volume={290},
  pages={116701},
  year={2021},
  publisher={Elsevier}
}

@article{xin2023model,
  title={Model Predictive Path Planning of AGVs: Mixed Logical Dynamical Formulation and Distributed Coordination},
  author={Xin, Jianbin and Wu, Xuwen and D’Ariano, Andrea and Negenborn, Rudy and Zhang, Fangfang},
  journal=IEEE_J_ITS,
  year={2023},
  publisher={IEEE}
}

@inproceedings{joshi2013distributed,
  title={Distributed SINR balancing for MISO downlink systems via the alternating direction method of multipliers},
  author={Joshi, S and Codreanu, Marian and Latva-aho, Matti},
  booktitle={Proc. 11th Int. Symp. Modeling Optim. Mobile},
  pages={318--325},
  year={2013},
  organization={IEEE}
}

@article{forero2010consensus,
  title={Consensus-based distributed linear support vector machines},
  author={Forero, Pedro A and Cano, Alfonso and Giannakis, Georgios B},
  journal={J. Mach. Learn. Res.},
  volume={11},
  pages={1663–1707},
  year={May 2010}
}

@article{schubiger2020gpu,
  title={GPU acceleration of ADMM for large-scale quadratic programming},
  author={Schubiger, Michel and Banjac, Goran and Lygeros, John},
  journal={J. Parallel Distrib. Comput.},
  volume={144},
  pages={55--67},
  year={2020},
  publisher={Elsevier}
}

@inproceedings{rostami2017admm,
  title={ADMM-based distributed model predictive control: Primal and dual approaches},
  author={Rostami, Ramin and Costantini, Giuliano and G{\"o}rges, Daniel},
  booktitle={Proc. IEEE Conf. Decis. Control},
  pages={6598--6603},
  year={2017},
  organization={IEEE}
}

@article{ghadimi2014optimal,
  title={Optimal parameter selection for the alternating direction method of multipliers (ADMM): Quadratic problems},
  author={Ghadimi, Euhanna and Teixeira, Andr{\'e} and Shames, Iman and Johansson, Mikael},
  journal=IEEE_J_AC,
  volume={60},
  number={3},
  pages={644--658},
  year={2014},
  publisher={IEEE}
}

@ARTICLE{9632418,
  author={Krupa, Pablo and Alvarado, Ignacio and Limon, Daniel and Alamo, Teodoro},
  journal=IEEE_J_CST, 
  title={Implementation of Model Predictive Control for Tracking in Embedded Systems Using a Sparse Extended ADMM Algorithm}, 
  year={2022},
  volume={30},
  number={4},
  pages={1798-1805},
  doi={10.1109/TCST.2021.3128824}}

@ARTICLE{7572966,
  author={Zheng, Huarong and Negenborn, Rudy R. and Lodewijks, Gabriël},
  journal=IEEE_J_CST, 
  title={Fast ADMM for Distributed Model Predictive Control of Cooperative Waterborne AGVs}, 
  year={2017},
  volume={25},
  number={4},
  pages={1406-1413},
  doi={10.1109/TCST.2016.2599485}}

@ARTICLE{6199971,
  author={Zheng, Yi and Li, Shaoyuan and Qiu, Hai},
  journal=IEEE_J_CST, 
  title={Networked Coordination-Based Distributed Model Predictive Control for Large-Scale System}, 
  year={2013},
  volume={21},
  number={3},
  pages={991-998},
  doi={10.1109/TCST.2012.2196280}}

@ARTICLE{8802250,
  author={Dunham, William and Hencey, Brandon and Girard, Anouck R. and Kolmanovsky, Ilya},
  journal=IEEE_J_CST, 
  title={Distributed Model Predictive Control for More Electric Aircraft Subsystems Operating at Multiple Time Scales}, 
  year={2020},
  volume={28},
  number={6},
  pages={2177-2190},
  doi={10.1109/TCST.2019.2932654}}

@ARTICLE{7546918,
  author={Zheng, Yang and Li, Shengbo Eben and Li, Keqiang and Borrelli, Francesco and Hedrick, J. Karl},
  journal=IEEE_J_CST, 
  title={Distributed Model Predictive Control for Heterogeneous Vehicle Platoons Under Unidirectional Topologies}, 
  year={2017},
  volume={25},
  number={3},
  pages={899-910},
  doi={10.1109/TCST.2016.2594588}}

@ARTICLE{10125035,
  author={Chiou, Suh-Wen},
  journal=IEEE_J_VT, 
  title={A Cooperative Agent-Based Traffic Signal Control for Vehicular Networks Under Stochastic Flow}, 
  year={2023},
  volume={72},
  number={10},
  pages={12592-12601},
  doi={10.1109/TVT.2023.3275208}}

@ARTICLE{pan2024asynchronous,
  author={Pan, ZhouDan and Cannon, Mark},
  journal={IEEE Trans. Control Netw. Syst.}, 
  title={Asynchronous ADMM via a Data Exchange Server}, 
  year={2024},
  volume={},
  number={},
  pages={1-12},
  doi={10.1109/TCNS.2024.3354840}}

@ARTICLE{zhong2024learning,
  author={Zhong, Jiaming and Mehrizi, Reza Valiollahi and Pirani, Mohammad and Yu, Chao and Kasaiezadeh, Alireza and Pant, Yash Vardhan and Khajepour, Amir},
  journal=IEEE_J_ITS, 
  title={Learning Agent-Based Model Predictive Control for Holistic Vehicle Performance}, 
  year={2024},
  volume={},
  number={},
  pages={1-11},
  doi={10.1109/TITS.2024.3435551}}

@article{liu2024distributed,
  title={Distributed model predictive control for virtually coupled heterogeneous trains: Comparison and assessment},
  author={Liu, Xiaoyu and Dabiri, Azita and Wang, Yihui and Xun, Jing and De Schutter, Bart},
  journal={IEEE Transactions on Intelligent Transportation Systems},
  year={2024},
  publisher={IEEE}
}

@article{zhang2025efficient,
  title={Efficient Federated Connected Electric Vehicle Scheduling System: A Noncooperative Online Incentive Approach},
  author={Zhang, Shiyao and Zhang, Shengyu},
  journal={IEEE Transactions on Intelligent Transportation Systems},
  year={2025},
  publisher={IEEE}
}

@article{khalatbarisoltani2023integrating,
  title={Integrating model predictive control with federated reinforcement learning for decentralized energy management of fuel cell vehicles},
  author={Khalatbarisoltani, Arash and Boulon, Loic and Hu, Xiaosong},
  journal={IEEE Transactions on Intelligent Transportation Systems},
  volume={24},
  number={12},
  pages={13639--13653},
  year={2023},
  publisher={IEEE}
}

@article{chen2016direct,
  title={The direct extension of ADMM for multi-block convex minimization problems is not necessarily convergent},
  author={Chen, Caihua and He, Bingsheng and Ye, Yinyu and Yuan, Xiaoming},
  journal={Mathematical Programming},
  volume={155},
  number={1},
  pages={57--79},
  year={2016},
  publisher={Springer}
}

@article{hong2017linear,
  title={On the linear convergence of the alternating direction method of multipliers},
  author={Hong, Mingyi and Luo, Zhi-Quan},
  journal={Mathematical Programming},
  volume={162},
  number={1},
  pages={165--199},
  year={2017},
  publisher={Springer}
}

@STRING{IEEE_J_ITS        = "{IEEE} Trans. Intell. Transp. Syst."}

@STRING{IEEE_J_VT         = "{IEEE} Trans. Veh. Technol."}

@STRING{IEEE_J_AC         = "{IEEE} Trans. Autom. Control"}

@STRING{IEEE_J_CST        = "{IEEE} Trans. Control Syst. Technol."}


\begin{IEEEbiography}[{\includegraphics[width=1in,height=1.25in, clip,keepaspectratio]{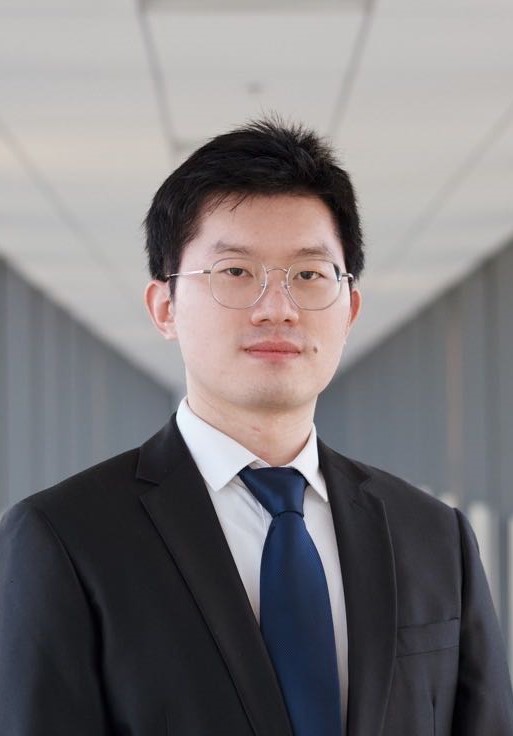}}]{Jiaming Zhong} (Student Member, IEEE) is currently a Ph.D. candidate at the University of Waterloo Mechatronic Vehicle Systems (MVS) Lab. He is also a co-founder and the lead of planning and control of LoopX Innovation Inc. in Ontario, Canada. He received his B.S. and MASc degrees in mechanical engineering from the Beijing Institute of Technology, China, in 2014 and 2017. 
He previously worked as a senior software engineer in SAIC Motor Co., Ltd. and NIO Co., Ltd. 
His research interests include learning-based planning and control, multi-agent theory, and autonomous driving.
\end{IEEEbiography}

\begin{IEEEbiography}[{\includegraphics[width=1in,height=1.25in, clip,keepaspectratio]{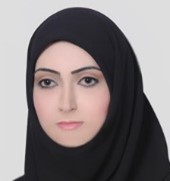}}]{Ladan Khoshnevisan} received her B.S. degree from the University of Science and Technology, Tehran, Iran in 2008, and her M.Sc. and Ph.D. degrees from Tarbiat Modares University and University of Tehran, Tehran, Iran in 2010 and 2017, respectively, all in electrical engineering. In 2017, she was an International Visiting Ph.D. Student in the Department of Applied Mathematics at the University of Waterloo. Subsequently, she served as a research assistant in the Department of Mechanical and Mechatronics Engineering at the University of Waterloo in 2022. From 2022 to 2024, she was a Postdoctoral Fellow in the Department of Applied Mathematics at the same university. She is currently a Postdoctoral Fellow in the Department of Mechanical and Mechatronics Engineering at the University of Waterloo, Waterloo, ON, Canada. Her research interests encompass robust and adaptive control systems, cyber security, automotive control, congestion and flow control in communication networks, bifurcation and system analysis, fault detection and isolation, fault-tolerant control, and fractional order systems.
\end{IEEEbiography}

\begin{IEEEbiography}[{\includegraphics[width=1in,height=1.25in, clip,keepaspectratio]{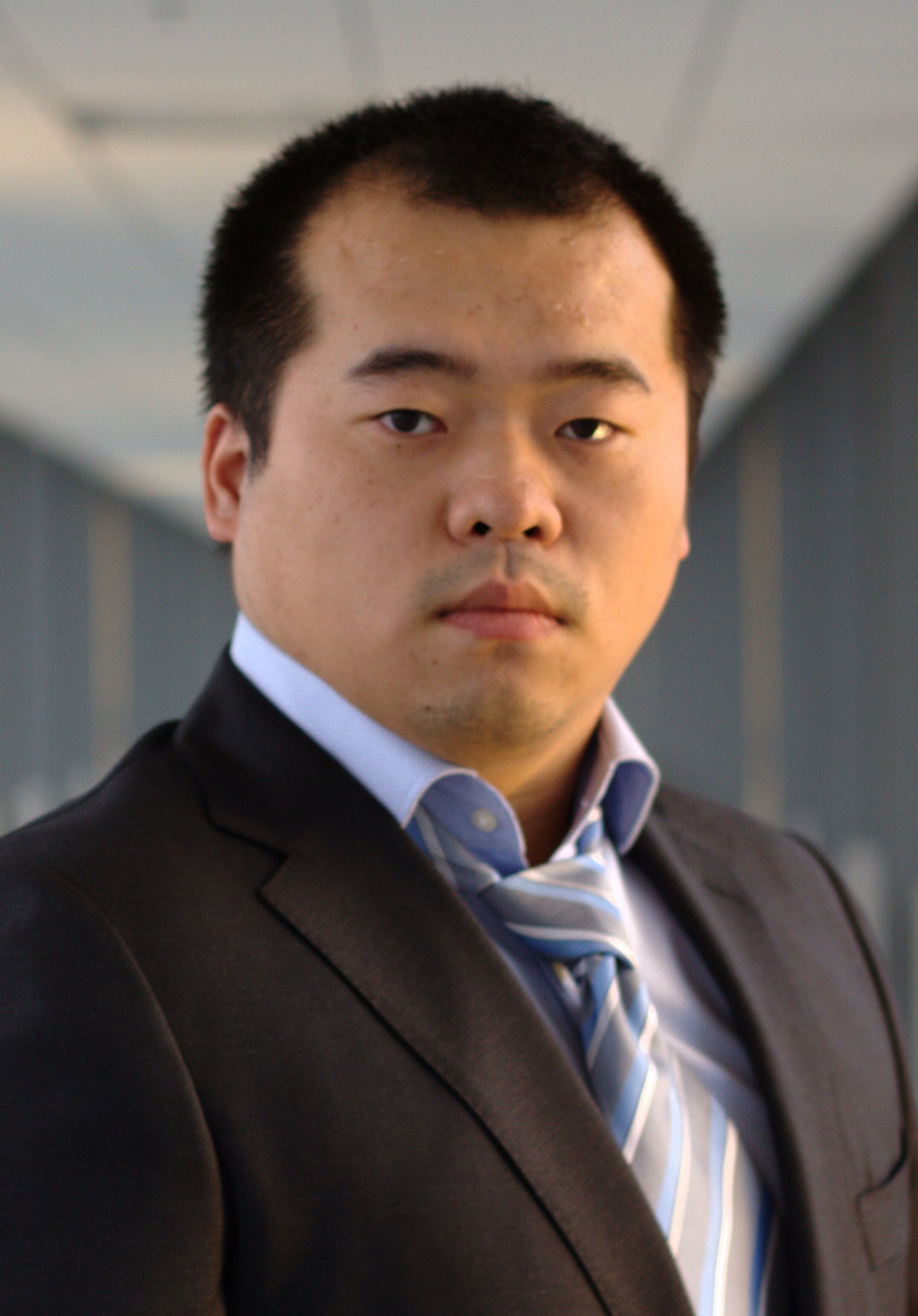}}]{Shucheng Huang} (Student Member, IEEE) received the B.S. degree in mechanical engineering from Pennsylvania State University, State College, USA, in 2018, and the MASc degree in mechanical and mechatronics engineering from the University of Waterloo, Waterloo, Canada, in 2020. He is currently working toward a Ph.D. degree at the University of Waterloo Mechatronic Vehicle Systems (MVS) Lab. His research interests include muti-modality LLM, muti-sensor fusion, and learning-based planning and control.
\end{IEEEbiography}

\begin{IEEEbiography}[{\includegraphics[width=1in,height=1.25in, clip,keepaspectratio]{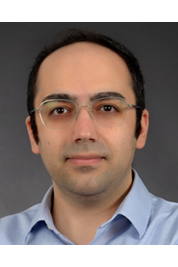}}]{Mohammad Pirani} (Senior Member, IEEE) is an assistant professor with the Department of Mechanical Engineering, University of Ottawa, Canada. He was a research assistant professor in the Department of Mechanical and Mechatronics Engineering at the University of Waterloo (2022–2023). He held postdoctoral researcher positions at the University of Toronto (2019–2021) and KTH Royal Institute of Technology, Sweden (2018–2019). He received a MASc degree in electrical and computer engineering and a Ph.D. degree in Mechanical and Mechatronics Engineering, both from the University of Waterloo in 2014 and 2017, respectively. His research interests include resilient and fault-tolerant control, networked control systems, and multi-agent systems.
\end{IEEEbiography}

\begin{IEEEbiography}[{\includegraphics[width=1in,height=1.25in, clip,keepaspectratio]{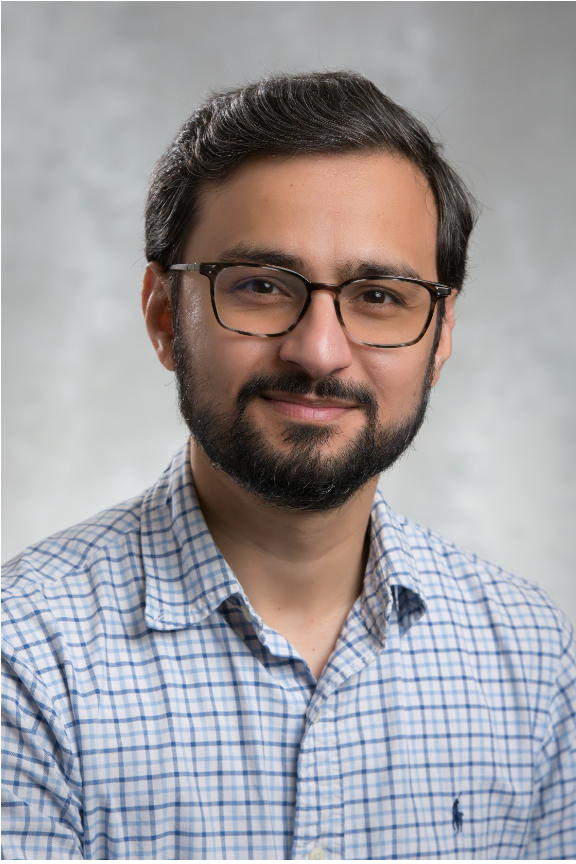}}]{Yash Vardhan Pant} Yash Vardhan Pant is an Assistant Professor in the Department of Electrical and Computer Engineering at the University of Waterloo. His research focuses on building robust and reliable autonomous systems using elements of Control Theory, Formal Methods, Machine Learning, and Optimization. He received a PhD in Electrical Engineering from the University of Pennsylvania in 2019, where he was a recipient of the Richard K. Dentel Memorial Award for research in Urban transportation. Prior to joining Waterloo in July 2021, he was a postdoctoral fellow at the Department of Electrical Engineering and Computer Sciences at the University of California, Berkeley.
\end{IEEEbiography}


\begin{IEEEbiography}[{\includegraphics[width=1in,height=1.25in, clip,keepaspectratio]{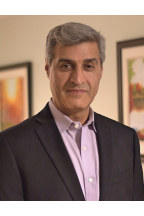}}]{Amir Khajepour} (Senior Member, IEEE) is a professor of Mechanical and Mechatronics Engineering and the Director of the Mechatronic Vehicle Systems (MVS) Lab at the University of Waterloo. He held the Tier 1 Canada Research Chair in Mechatronic Vehicle Systems from 2008 to 2022 and the Senior NSERC/General Motors Industrial Research Chair in Holistic Vehicle Control from 2017 to 2022. His work has led to training over 150 PhD and MASc students, filling 30 patents, publishing 600 research papers, numerous technology transfers, and establishing several start-up companies. He has been recognized with the Engineering Medal from Professional Engineering Ontario and is a fellow of the Engineering Institute of Canada, the American Society of Mechanical Engineering, and the Canadian Society of Mechanical Engineering.
\end{IEEEbiography}




\end{document}